\documentclass[acmsmall,screen,]{acmart}
\pdfoutput=1 
\setcopyright{none}

\AtBeginDocument{%
  }

\acmJournal{CSUR}

\usepackage{microtype}
\usepackage{float}
\usepackage{multirow}
\usepackage{subcaption}
\usepackage[normalem]{ulem}
\usepackage{enumitem}
\usepackage{xcolor}
\usepackage{soul}
\colorlet{soulred}{red!30}
\sethlcolor{soulred}%
\usepackage{framed}
\usepackage{url}
\usepackage{pifont}
\usepackage{colortbl}
\usepackage{stfloats}
\usepackage{amssymb}
\usepackage{epigraph}
\usepackage{hyperref}

\definecolor{green}{rgb}{0.1,0.1,0.1}
\definecolor{chocolate}{HTML}{D2691E}
\definecolor{maroon}{HTML}{A00000}
\definecolor{indigo}{HTML}{4B0082}
\definecolor{green}{HTML}{008000}
\definecolor{cadmiumgreen}{rgb}{0.0, 0.42, 0.24}

\definecolor{airforceblue}{rgb}{0.36, 0.54, 0.66}
\definecolor{Gray}{gray}{0.9}

\usepackage[edges]{forest}
\definecolor{lightcoral}{rgb}{0.94, 0.5, 0.5}
\definecolor{lightgreen}{rgb}{0.56, 0.93, 0.56}

\definecolor{brightlavender}{rgb}{0.75, 0.58, 0.89}
\definecolor{capri}{rgb}{0.0, 0.75, 1.0}
\definecolor{carminepink}{rgb}{0.92, 0.3, 0.26}
\definecolor{celadon}{rgb}{0.67, 0.88, 0.69}
\definecolor{darkpastelgreen}{rgb}{0.01, 0.75, 0.24}

\definecolor{pastelblue}{rgb}{0.68, 0.78, 0.81}
\definecolor{mintgreen}{rgb}{0.6, 0.98, 0.6}
\definecolor{lavender}{rgb}{0.71, 0.49, 0.86}
\definecolor{peach}{rgb}{1.0, 0.9, 0.71}
\definecolor{coral}{rgb}{1.0, 0.5, 0.31}
\definecolor{mauve}{rgb}{0.88, 0.69, 1.0}
\definecolor{lemonyellow}{rgb}{1.0, 0.96, 0.4}

\definecolor{hidden-draw}{RGB}{205, 44, 36}
\definecolor{hidden-blue}{RGB}{194,232,247}
\definecolor{hidden-orange}{RGB}{243,202,120}
\definecolor{hidden-yellow}{RGB}{242,244,193}
\definecolor{tree-level-1}{RGB}{245,20,85}
\definecolor{tree-level-2}{RGB}{246,86,118}
\definecolor{tree-level-3}{RGB}{248,177,193}
\definecolor{tree-leaf}{RGB}{176,230,198}

\definecolor{Self}{RGB}{255,0,128}
\definecolor{Ensemble}{RGB}{0,127,255}
\definecolor{Iterative}{RGB}{153,51,255}

\definecolor{exemplar1}{RGB}{136,98,148}
\definecolor{exemplar2}{RGB}{148,210,242}
\definecolor{knowledge1}{RGB}{249,219,152}
\definecolor{knowledge2}{RGB}{255,245,220}

\definecolor{lighttealblue}{RGB}{41, 157, 143}
\definecolor{lightplum}{RGB}{233, 196, 106}
\definecolor{harvestgold}{RGB}{216, 118, 89}

\begin{document}


\title{LLM4SR: A Survey on Large Language Models for Scientific Research}

\author{Ziming Luo}
\authornote{Both authors contributed equally to this work.}
\email{ziming.luo@utdallas.edu}
\orcid{0009-0005-0310-677X}
\affiliation{%
 \institution{University of Texas at Dallas}
  \city{Dallas}
  \state{Texas}
  \country{USA}
}

\author{Zonglin Yang}
\email{zonglin001@ntu.edu.sg}
\authornotemark[1]
\orcid{0000-0002-8059-6654}
\affiliation{%
  \institution{Nanyang Technological University}
  \city{Singapore}
  \country{Singapore}
}

\author{Zexin Xu}
\email{zexin.xu@utdallas.edu}
\orcid{0009-0002-2065-336X}
\affiliation{%
  \institution{University of Texas at Dallas}
  \city{Dallas}
  \state{Texas}
  \country{USA}
}

\author{Wei Yang}
\email{wei.yang@utdallas.edu}
\orcid{0000-0002-5338-7347}
\affiliation{%
  \institution{University of Texas at Dallas}
  \city{Dallas}
  \state{Texas}
  \country{USA}
}

\author{Xinya Du}
\email{xinya.du@utdallas.edu}
\orcid{0000-0003-4255-8013}
\affiliation{%
 \institution{University of Texas at Dallas}
 \city{Dallas}
 \state{Texas}
 \country{USA}}

\renewcommand{\shortauthors}{Luo and Yang et al.}

\begin{abstract}
In recent years, the rapid advancement of Large Language Models (LLMs) has transformed the landscape of scientific research, offering unprecedented support across various stages of the research cycle. 
This paper presents the first systematic survey dedicated to exploring how LLMs are revolutionizing the scientific research process. We analyze the unique roles LLMs play across four critical stages of research: hypothesis discovery, experiment planning and implementation, scientific writing, and peer reviewing. Our review comprehensively showcases the task-specific methodologies and
evaluation benchmarks. 
By identifying current challenges and proposing future research directions, this survey not only highlights the transformative potential of LLMs, but also aims to inspire and guide researchers and practitioners in leveraging LLMs to advance scientific inquiry. 
Resources are available at the following
repository: \url{https://github.com/du-nlp-lab/LLM4SR}.

\end{abstract}



\maketitle

\begin{figure}[h!]
    \centering
    \resizebox{\textwidth}{!}{
    \includegraphics[]{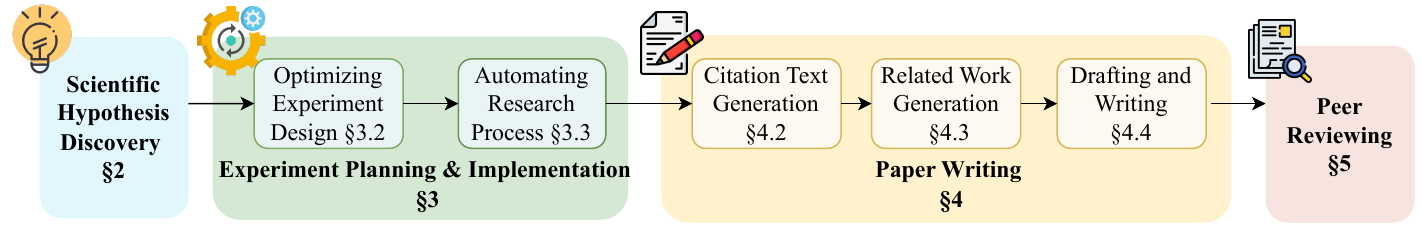} 
    }
    \caption{Schematic overview of the scientific research pipeline covered in this survey. This cyclical process begins with scientific hypothesis discovery, followed by experiment planning and implementation, paper writing, and finally peer reviewing of papers. The experiment planning stage consists of optimizing experiment design and executing research tasks, while the paper writing stage consists of citation text generation, related work generation, and drafting \& writing.}
    \Description{Schematic overview of the scientific research pipeline covered in this survey. This cyclical process begins with scientific hypothesis discovery, followed by experiment planning and implementation, paper writing, and finally peer reviewing of papers. The experiment planning stage consists of optimizing experiment design and executing research tasks, while the paper writing stage consists of citation text generation, related work generation, and drafting \& writing.}
    \label{fig:schematic_overview}
\end{figure}

\newpage


\section{Introduction}\label{sec:intro}

\epigraph{\textit{``If I have seen further, it is by standing on the shoulders of giants.''}}{--- \textit{Isaac Newton}}

The scientific research pipeline is a testament to the achievements of the Enlightenment in systematic inquiry~\citep{chalmers2013thing, jevons1877principles, jevons1877principles}. In this traditional paradigm, scientific research involves a series of well-defined steps: researchers start by gathering background knowledge, propose hypotheses, design and execute experiments, collect and analyze data, and finally report findings through a manuscript that undergoes peer review. This cyclical process has led to groundbreaking advancements in modern science and technology, yet it remains constrained by the creativity, expertise, and finite time and resources available inherent to human researchers. 

For decades, the scientific community has sought to enhance this process by automating aspects of scientific research, aiming to increase the productivity of scientists.
Early computer-assisted research can date back to the 1970s, introducing systems such as Automated Mathematician~\citep{lenat1977automated, lenat1984and} and BACON~\citep{Bacon1977}, which showed the potential of machines to assist in specialized research tasks like theorem generation and empirical law identification. 
More recently, systems such as AlphaFold~\citep{jumper2021highly} and OpenFold~\citep{ahdritz2024openfold} have exemplified pioneering efforts to automate specific research tasks, significantly speeding up scientific progress in their respective domains by thousands of times.
Yet it was only with the advent of foundation models and the recent explosion in Large Language Models (LLMs)~\citep{Touvron2023LLaMAOA, achiam2023gpt} that the vision of comprehensive AI assistance across multiple research domains became realistic~\citep{survey2023zhao}.

The recent years have witnessed remarkable advancements in LLMs, transforming various fields of AI and Natural Language Processing (NLP).
These models, such as GPT-4~\citep{achiam2023gpt} and LLaMA~\citep{Touvron2023LLaMAOA}, have set new benchmarks in understanding, generating and interacting with human language. Their capabilities, enhanced by massive datasets and innovative architectures, now extend beyond conventional NLP tasks to more complex and domain-specific challenges. In particular, the ability of LLMs to process massive amounts of data, generate human-like text, and assist in complex decision-making has captured significant attention in the scientific community~\citep{liang2024mapping, DBLP:journals/corr/abs-2409-04109}. These breakthroughs suggest that LLMs have the potential to revolutionize the way scientific research is conducted, documented, and evaluated~\citep{wang2023scientific, DBLP:conf/acl/YangDLZPC24, wang2024autosurvey}.

\tikzstyle{my-box}=[
    rectangle,
    rounded corners,
    text opacity=1,
    minimum height=1.5em,
    minimum width=5em,
    inner sep=2pt,
    align=center,
    fill opacity=.5,
]
\tikzstyle{hypothesis_leaf}=[my-box, minimum height=1.5em,
    fill=cyan!20, text=black, align=left,font=\small,
    inner xsep=2pt,
    inner ysep=4pt,
]
\tikzstyle{cause_leaf}=[my-box, minimum height=1.5em,
    fill=lighttealblue!20, text=black, align=left,font=\small,
    inner xsep=2pt,
    inner ysep=4pt,
]
\tikzstyle{detect_leaf}=[my-box, minimum height=1.5em,
    fill=lightplum!20, text=black, align=left,font=\small,
    inner xsep=2pt,
    inner ysep=4pt,
]
\tikzstyle{mitigate_leaf}=[my-box, minimum height=1.5em,
    fill=harvestgold!20, text=black, align=left,font=\small,
    inner xsep=2pt,
    inner ysep=4pt,
]

\begin{figure}[!ht]
    \centering
    \resizebox{\textwidth}{!}{%
    \begin{forest}
        forked edges,
        for tree={
            grow=east,
            reversed=true,
            anchor=base west,
            parent anchor=east,
            child anchor=west,
            base=left,
            font=\normalsize,
            rectangle,
            rounded corners,
            align=left,
            minimum width=4em,
            edge+={darkgray, line width=1pt},
            s sep=3pt,
            inner xsep=2pt,
            inner ysep=3pt,
            ver/.style={rotate=90, child anchor=north, parent anchor=south, anchor=center},
        },
        where level=1{text width=6.0em,font=\normalsize}{},
        where level=2{text width=9.5em,font=\normalsize}{},
        where level=3{text width=8.0em,font=\small}{},
        where level=4{text width=7.5em,font=\small}{},
            [
                {Large Language Models (LLMs) for Scientific Research}, ver, color=carminepink!100, fill=carminepink!15, text=black
                [
                    Scientific \\ Hypothesis \\ Discovery \\(\S~\ref{section3}), color=cyan!100, fill=cyan!100, text=black
                    [
                        History (\S~\ref{appen:history_of_discovery}), color=cyan!100, fill=cyan!60, text=black
                        [
                            Literature-based \\ Discovery (\S~\ref{appen:history_lbd}), color=cyan!100, fill=cyan!40, text=black
                            [
                                { LBD~\citep{swanson1986undiscovered,DBLP:journals/cacm/HopeDWEH23}, DBLP~\citep{DBLP:journals/nature/TshitoyanDWDRKP19},
                                Link Prediction Models~\citep{DBLP:conf/acl/WangHJKJBL19, DBLP:conf/cikm/SybrandtTSS20, DBLP:conf/acl/XuSXFWZ23}}, hypothesis_leaf, text width=29.1em   
                            ]
                        ]
                        [
                            Inductive Reasoning \\ (\S~\ref{history_inductive}), color=cyan!100, fill=cyan!40, text=black
                            [
                                {\citet{norton2003little}, \citet{DBLP:journals/corr/abs-2303-12023}, \citet{DBLP:journals/corr/abs-2212-10923},             \citet{DBLP:conf/nips/ZhongZLAKS23}, \\ \citet{DBLP:journals/corr/abs-2310-07064},\citet{DBLP:journals/corr/abs-2309-05660}, \citet{DBLP:conf/iclr/QiuJLSPBWK0D024} }, hypothesis_leaf, text width=29.1em   
                            ] 
                        ]
                    ]
                    [
                        Development of \\ Methods (\S~\ref{sec:discovery_method}), color=cyan!100, fill=cyan!60, text=black
                        [
                            Main Trajectory \\ (\S~\ref{subsec:main_trajectory}), color=cyan!100, fill=cyan!40, text=black
                            [
                                { \textsc{SciMON}~\citep{DBLP:conf/acl/0005DJH24}, \textsc{MOOSE}~\citep{DBLP:conf/acl/YangDLZPC24}, \textsc{MCR}~\citep{DBLP:conf/emnlp/SprueillEOSJC23}, \textsc{Qi}~\citep{DBLP:journals/corr/abs-2311-05965},\\
                                \textsc{FunSearch}~\citep{DBLP:journals/nature/RomeraParedesBNBKDREWFKF24},   \textsc{ChemReasoner}~\citep{DBLP:conf/icml/SprueillEAOSJLJ24}, 
                                \textsc{HypoGeniC}~\citep{DBLP:journals/corr/abs-2404-04326},\\
                                \textsc{ResearchAgent}~\citep{DBLP:journals/corr/abs-2404-07738},
                                \textsc{LLM-SR}~\citep{DBLP:journals/corr/abs-2404-18400},
                                \textsc{SGA}~\citep{DBLP:conf/icml/MaWGSTRGM24},
                                \textsc{AIScientist}~\citep{lu2024ai}, \\
                                \textsc{MLR-Copilot}~\citep{li2024mlr},
                                \textsc{IGA}~\citep{DBLP:journals/corr/abs-2409-04109},
                                \textsc{SciAgents}~\citep{DBLP:journals/corr/abs-2409-05556},
                                \textsc{Scideator}~\citep{DBLP:journals/corr/abs-2409-14634},\\
                                \textsc{MOOSE-Chem}~\citep{yang2024moose}, 
                                \textsc{VirSci}~\citep{su2024two},
                                \textsc{CoI}~\citep{li2024chain},
                                \textsc{Nova}~\citep{hu2024nova},\\
                                \textsc{CycleResearcher}~\citep{weng2024cycleresearcher},
                                \textsc{SciPIP}~\citep{Wang2024SciPIPAL}
                                }, hypothesis_leaf, text width=29.1em   
                            ] 
                        ]
                        [
                            Other Methods \\ (\S~\ref{subsec:other_method}), color=cyan!100, fill=cyan!40, text=black
                            [
                                { \textsc{Socratic reasoning}\cite{DBLP:journals/corr/abs-2309-05689},
                                \textsc{IdeaSynth}~\citep{pu2024ideasynth},
                                \textsc{HypoRefine}~\citep{liu2024literature},
                                 \textsc{LDC}~\citep{li2024learning}
                                }, hypothesis_leaf, text width=29.1em   
                            ] 
                        ]
                    ]
                    [
                        Benchmarks (\S~\ref{sec:discovery_benchmark}), color=cyan!100, fill=cyan!60, text=black
                        [
                        { \textsc{SciMON}~\citep{DBLP:conf/acl/0005DJH24}, \textsc{Tomato}~\citep{DBLP:conf/acl/YangDLZPC24}, \citet{DBLP:journals/corr/abs-2311-05965}, \citet{kumar2024can}, \textsc{Tomato-Chem}~\citep{yang2024moose}\\
                        \textsc{DiscoveryBench}~\citep{DBLP:journals/corr/abs-2407-01725}, \textsc{DiscoveryWorld}~\citep{DBLP:journals/corr/abs-2406-06769}}, hypothesis_leaf, text width=38.7em   
                        ]
                    ]
                    [
                        Evaluation (\S~\ref{sec:discovery_evaluation}), color=cyan!100, fill=cyan!60, text=black
                        [
                        { LLM-based / Expert-based Evaluation; Direct Evaluation / Reference-based Evaluation; \\
                        Direct Evaluation / Comparison-based Evaluation; Real Experiment Evaluation}, hypothesis_leaf, text width=38.7em   
                        ]
                    ]
                ]
                [
                    Experiment \\ Planning \\ and Imple-\\ mentation \\(\S~\ref{section4}), color=lighttealblue!100, fill=lighttealblue!100, text=black
                    [
                        Optimizing Experi- \\mental Design (\S~\ref{section4.2}), color=lighttealblue!100, fill=lighttealblue!40, text=black
                        [
                            {
                            HuggingGPT~\citep{shen2024hugginggpt}, CRISPR-GPT~\citep{CRISPR-GPT}, ChemCrow~\citep{m2024augmenting}, Coscientist~\citep{boiko2023autonomous}, LLM-RDF~\citep{ruan2024automatic},\\ AutoGen~\citep{autogen2023},
                            \citet{liiqa}, \citet{li2024simulating}
                            }, cause_leaf, text width=38.7em   
                        ] 
                    ]
                    [
                        Automating Experi- \\mental Process (\S \ref{section4.3}), color=lighttealblue!100, fill=lighttealblue!60, text=black
                        [
                            Data Preparation \\ (\S~\ref{section4.3.1}), color=lighttealblue!100, fill=lighttealblue!40, text=black
                            [
                                { Clearning~\citep{zhang2023jellyfish, chen2024data}, Labeling~\citep{tan2024large},  Feature Engineering~\citep{hollmann2024large},\\Synthesis~\citep{Li2023AreYI, liu2023training, Li2024FinegrainedHD}}
                                , cause_leaf, text width=29.1em   
                            ] 
                        ]
                        [
                            Experiment Execution\\ and Workflow \\ Automation (\S~\ref{section4.3.2}), color=lighttealblue!100, fill=lighttealblue!40, text=black
                            [
                                {ChemCrow~\citep{m2024augmenting}, Coscientist~\citep{boiko2023autonomous}, \citet{Wang2024Efficient}, 
                                \citet{Ramos2023},\\ ChatDrug~\citep{Liu2024}, DrugAssist~\citep{DrugAssist},
                                ESM-1b~\citep{RivesMSGLLGOZMF21}, 
                                ESM-2~\citep{lin2023evolutionary}, \\
                                \citet{FerruzH22}, \citet{he2024novo}}
                                , cause_leaf, text width=29.1em   
                            ] 
                        ]
                        [
                            Data Analysis and \\ Interpretation (\S \ref{section4.3.3}), color=lighttealblue!100, fill=lighttealblue!40, text=black
                            [
                                { \citet{singh2024rethinking}, \citet{li2024automated}, \textsc{MentalLLaMA}\citep{yang2024mentallama}, \\\citet{dai2023llm}, \citet{rasheed2024can}, \citet{zhao2024revolutionizing}, \citet{oliver2024opening}}
                                , cause_leaf, text width=29.1em   
                            ] 
                        ]
                    ]
                    [
                        Benchmarks \& \\ Evaluation  (\S \ref{section4.4}), color=lighttealblue!100, fill=lighttealblue!60, text=black
                        [
                            { \textsc{TaskBench}~\citep{shen2023TaskBench}, \textsc{DiscoveryWorld}~\citep{DBLP:journals/corr/abs-2406-06769}, \textsc{MLAgentBench}~\citep{huang2024mlagentbenchevaluatinglanguageagents},  \textsc{AgentBench}~\citep{liu2023agentbenchevaluatingllmsagents},                            \textsc{Spider2-V}~\citep{cao2024spider2vfarmultimodalagents},\\
                            \textsc{DSBench}~\citep{jing2024dsbenchfardatascience},                            DS-1000 \citep{lai2022ds1000naturalreliablebenchmark}, \textsc{CORE-Bench}~\citep{siegel2024corebenchfosteringcredibilitypublished},
                            \textsc{SUPER}~\citep{bogin2024superevaluatingagentssetting},
                            \textsc{MLE-Bench}~\citep{chan2024mlebenchevaluatingmachinelearning},
                            \textsc{LAB-Bench}~\citep{Laurent2024LABBenchMC},\\
                            \textsc{ScienceAgentBench}~\citep{chen2024scienceagentbench}
                            }, cause_leaf, text width=38.7em
                        ]
                    ]
                ]
                [                    
                    Paper Writing \\ (\S~\ref{section5}), color=lightplum!100, fill=lightplum!100, text=black
                    [
                        Citation Text \\ Generation  (\S~\ref{section5.2}), color=lightplum!100, fill=lightplum!60, text=black
                        [
                            { \citet{xing-etal-2020-automatic}, \textsc{AutoCite}~\citep{Wang2023AutoCite}, \textsc{BACO}~\citep{Ge2021BACOAB}, \citet{gu-hahnloser-2024-controllable}, \citet{math10101763}}
                            , detect_leaf, text width=38.7em
                        ]
                    ]
                    [
                        Related Work \\ Generation (\S~\ref{section5.3}), color=lightplum!100, fill=lightplum!60, text=black
                        [
                            { \citet{Zimmermann2024Leveraging}, \citet{shi2023towards, Agarwal2024LitLLMAT, hu2024hireviewhierarchicaltaxonomydrivenautomatic, Yu2024Reinforced}, \\ \citet{ susnjak2024automating}, \textsc{LitLLM}~\citep{Agarwal2024LitLLMAT}, \textsc{HiReview}~\citep{hu2024hireviewhierarchicaltaxonomydrivenautomatic}, \citet{nishimura-etal-2024-toward}}
                            , detect_leaf, text width=38.7em
                        ]
                    ]
                    [
                        Drafting and Writing \\ (\S~\ref{section5.4}), color=lightplum!100, fill=lightplum!60, text=black
                        [
                            { \citet{august-etal-2022-generating}, \textsc{SCICAP}~\citep{hsu2021scicap}, \textsc{PaperRobot}~\citep{DBLP:conf/acl/WangHJKJBL19}, \citet{ifargan2024autonomous}, \textsc{CoAuthor}~\citep{Lee2022CoAuthor}, \\ \textsc{AutoSurvey}~\citep{wang2024autosurvey}, \textsc{AI Scientist}~\citep{lu2024ai} }
                            , detect_leaf, text width=38.7em
                        ]
                    ]
                    [
                        Benchmarks \& \\ Evaluation (\S \ref{section5.5}), color=lightplum!100, fill=lightplum!60, text=black
                        [
                            { ALCE~\citep{gao-etal-2023-enabling}, \textsc{CiteBench}~\citep{Funkquist2022CiteBenchAB}, \textsc{SciGen}~\citep{moosavi2021scigen}, \textsc{SciXGen}~\citep{chen-etal-2021-scixgen-scientific}}
                            , detect_leaf, text width=38.7em
                        ]
                    ]
               ]
               [
                    Peer Review- \\ing (\S~\ref{section6}), color=harvestgold!100, fill=harvestgold!100, text=black
                    [
                        Automated Peer \\ Reviewing Generation \\(\S~\ref{section6.2.1}), color=harvestgold!100, fill=harvestgold!60, text=black
                        [
                            { \textsc{ReviewRobot}~\citep{Wang_2020_ReviewRobot}, \textsc{Reviewer2}~\citep{gao2024reviewer2optimizingreviewgeneration}, \
                            \textsc{SWIF2T}~\citep{Chamoun_2024_AutomatedFocusedFeedback}, 
                            \textsc{SEA}~\citep{Yu_2024_AutomatedPeerReviewing}, \textsc{MARG}~\citep{DArcy_2024_MARG}, 
                            \textsc{MetaGen}~\citep{Bhatia2020MetaGenAA}, \\ Kumar et al. \citep{Kumar2021ADN}, \textsc{MReD}~\citep{Chenhui_2022_MReDDataset}, \textsc{CGI2}~\citep{zeng_2024_scientificopinionsummarizationpaper}, \textsc{CycleReviewer}~\citep{weng2024cycleresearcher}},
                            mitigate_leaf, text width=38.7em
                        ]
                    ]
                    [
                        LLM-assisted Peer \\ Review Workflows \\ (\S~\ref{section6.2.2}), color=harvestgold!100, fill=harvestgold!60, text=black
                        [
                            Information \\ Summarization, color=harvestgold!100, fill=harvestgold!20, text=black
                            [
                                {\textsc{PaperMage}~\citep{Lo_2023_PaperMageToolkit}, \textsc{CocoSciSum}~\citep{Ding_2023_CocoSciSum}},
                                mitigate_leaf, text width=29.1em
                            ]
                        ]
                        [
                            Error Detection \& \\ Quality Verification, color=harvestgold!100, fill=harvestgold!20, text=black
                            [
                                {\textsc{ReviewerGPT}~\citep{Liu_2023_ReviewerGPT}, \textsc{PaperQA2}~\citep{Skarlinski_2024_LanguageAgentsSynthesis}, \textsc{Scideator}~\citep{Radensky2024ScideatorHS}},
                                mitigate_leaf, text width=29.1em
                            ]
                        ]
                        [
                            Review Writing \\ Support, color=harvestgold!100, fill=harvestgold!20, text=black
                            [
                                {\textsc{ReviewFlow}~\citep{Sun_2024_ReviewFlowScaffolding}, CARE~\citep{Zyska_2023_CAREEnvironment}, \textsc{DocPilot}~\citep{Mathur2024DocPilotCF}},
                                mitigate_leaf, text width=29.1em
                            ]
                        ]
                    ]
                    [
                        Benchmarks \& \\ Evaluation (\S~\ref{section6.3}), color=harvestgold!100, fill=harvestgold!60, text=black
                        [
                            { MOPRD~\citep{Lin_2023_MOPRD}, ORSUM~\citep{zeng_2024_scientificopinionsummarizationpaper}, 
                            \textsc{MReD}~\citep{Chenhui_2022_MReDDataset}, 
                            \textsc{PeerSum}~\citep{Li_2023_SummarizingMultipleDocuments}, \textsc{NLPeer}~\citep{Nils_2022_NLPeerResource}, 
                            \textsc{PeerRead}~\citep{Kang_2018_PeerReadDataset}, \\
                            \textsc{ASAP-Review}~\citep{yuan2021automatescientificreviewing}, \textsc{ReviewCritique}~\citep{Du_2024_LLMsAssistNLPResearchers}, \textsc{Reviewer2}~\citep{gao2024reviewer2optimizingreviewgeneration}}, mitigate_leaf, text width=38.7em
                        ]
                    ]
                ]
            ]
    \end{forest}
    }
    \Description{The main content flow and categorization of this survey.}
    \caption{The main content flow and categorization of this survey.}
    \label{fig:categorization_of_survey}
\end{figure}
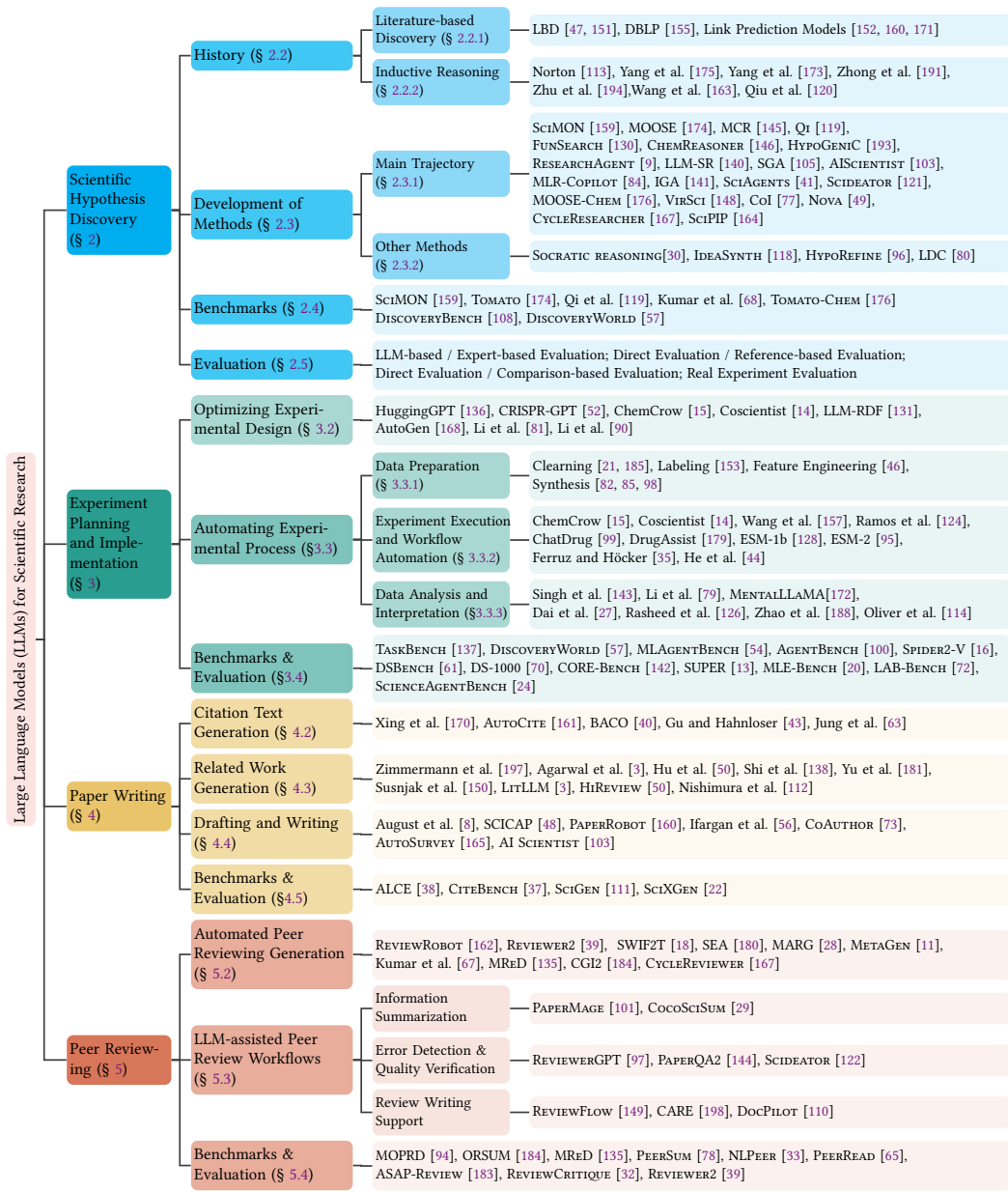

In this survey, we explore how LLMs are currently being applied across various stages of the scientific research process. Specifically, we identify four general tasks where LLMs have demonstrated notable potential. We begin by exploring their application in scientific hypothesis discovery, where LLMs leverage existing knowledge and experimental observations to suggest novel research ideas. 
This is followed by a review of their contributions to experiment planning and implementation, where LLMs aid in optimizing experimental design, automating workflows, and analyzing data. We also cover their use in scientific writing, including the generation of citations, related work sections, and even drafting entire papers. 
Finally, we discuss their potential in peer review, where LLMs support the evaluation of scientific papers by offering automated reviews and identifying errors or inconsistencies.
For each of these tasks, we provide a comprehensive review of the methodologies, benchmarks, and evaluation methods.
Moreover, the survey identifies the limitations of each task and highlights areas needing improvement.
By analyzing the various stages of the research cycle where LLMs contribute, this survey can inspire researchers to explore emerging concepts, develop evaluation metrics, and design innovative approaches to integrate LLMs into their workflows effectively.

{\bf Comparison with Existing Surveys.} 
This survey provides a broader and more comprehensive perspective on the applications of LLMs across the entire scientific research cycle compared to prior specialized studies. For example, \citet{zhang2024comprehensive} review over 260 LLMs in scientific discovery across various disciplines, focusing primarily on technical aspects such as model architectures and datasets, without situating their roles within the broader context of the research process. Similarly, other surveys tend to adopt narrower scopes, examining specific capabilities of LLMs for general applications, such as planning~\citep{huang2024understanding} or automation~\citep{wang2024survey}, rather than their focused utility in scientific research workflows.
Additionally, some works address general approaches relevant to specific research stages but are not exclusively centered on LLMs, such as related work and citation text generation~\citep{Li2024RelatedWA} or peer review processes~\citep{Nils_2022_NLPeerResource}.
In contrast, this survey integrates these fragmented perspectives, providing a holistic analysis of LLMs' contributions across the scientific workflow and highlighting their potential to address the diverse and evolving demands of modern research.

{\bf Organization of this Survey.} 
As illustrated in Figure~\ref{fig:categorization_of_survey}, the structure of this survey is as follows: \S~\ref{section3} covers LLMs for scientific hypothesis discovery, including an overview of methodologies and key challenges. \S~\ref{section4} focuses on experiment planning and implementation, highlighting how LLMs can optimize and automate these processes. \S~\ref{section5} delves into automated paper writing, including citation and related work generation, while \S~\ref{section6} explores LLM-assisted peer review. For each topic, the survey concludes with a summary of current challenges and future directions in this rapidly evolving field.




\section{LLMs for Scientific Hypothesis Discovery}
\label{section3}

\subsection{Overview}

%
Before the emergence of the field ``LLMs for scientific hypothesis discovery'', the most related previous research domains are ``literature-based discovery'' and ``inductive reasoning''.
We first summarize the research in the two related domains~(as history), then summarize the methods, benchmarks, evaluation development trends, and important progress, and finally conclude with the main challenges in the discovery task.


\subsection{History of Scientific Discovery}
\label{appen:history_of_discovery}

Using LLMs to generate novel scientific hypotheses is a new research topic, mostly originating from two related research domains, which are ``literature-based discovery'' and ``inductive reasoning''. 

\subsubsection{Literature-based Discovery}
\label{appen:history_lbd}
Literature-based discovery~(LBD) was first proposed by \citet{swanson1986undiscovered}.
The central idea is that ``knowledge can be public, yet undiscovered, if independently created fragments are logically related but never retrieved, brought together, and interpreted.''
Therefore, how to retrieve public knowledge that can be brought together to create new knowledge remains a challenge.

\citet{swanson1986undiscovered} propose a classic formalization of LBD, which is the “ABC” model where two concepts A and C are hypothesized as linked if they both co-occur with some intermediate concept B in papers.
More recent work has used word vectors~\citep{DBLP:journals/nature/TshitoyanDWDRKP19} or link prediction models~\citep{DBLP:conf/acl/WangHJKJBL19,DBLP:conf/cikm/SybrandtTSS20,DBLP:conf/acl/XuSXFWZ23} to discover links between concepts to compose hypotheses.

However, classic LBD methods do not model contexts that human scientists
consider in the ideation process, and are limited to predicting pairwise relations between discrete concepts~\citep{DBLP:journals/cacm/HopeDWEH23}.
To overcome these limitations, \citet{DBLP:conf/acl/0005DJH24} make the first attempt to ground LBD in a natural language context to constrain the generation
space, and also use generated sentences as output instead of only predicting relations as in the traditional LBD.

Another limitation of LBD is that it has long been thought of as only be applicable to a very specific, narrow type of hypothesis~\citep{DBLP:conf/acl/0005DJH24}.
However, recent progress in scientific discovery indicates that LBD might have a much wider applicable scope.
Particularly, \citet{DBLP:conf/acl/YangDLZPC24} and \citet{yang2024moose} discuss extensively with social science and chemistry researchers correspondingly, and find that most existing social science and chemistry published hypotheses~(instead of only a narrow type of hypotheses) can be formulated in a LBD pattern.
It probably indicates that future hypotheses in social science and chemistry to be published can also result from (correct) linkages and associations of existing knowledge.

\subsubsection{Inductive Reasoning}
\label{history_inductive}
Inductive reasoning is about finding a general ``rule'' or ``hypothesis'' that has a wide application scope from specific ``observations''~\citep{DBLP:journals/corr/abs-2303-12023}. 
For example, Geocentrism, Heliocentricism, and Newton's Law of Gravity are all proposed ``rules'' based on the ``observations'' of the movements of stars and planets.
Scientific discovery is a difficult task of inductive reasoning to an extreme, where each ``rule'' is a novel scientific finding.

The philosophy of science community has summarized three fundamental requirements for a ``rule'' from inductive reasoning~\citep{norton2003little}, which are
(1) ``rule'' should not be in conflict with ``observations'';
(2) ``rule'' should reflect the reality;
(3) ``rule'' should present a general pattern that can be applied to a larger scope than the ``specific'' observations, covering new information not existing in the observations.
Previously inductive reasoning research is mainly conducted by the ``inductive logic programming'' community~\citep{DBLP:journals/jair/CropperD22}, which uses formal language and symbolic reasoners.
\citet{DBLP:journals/corr/abs-2212-10923} first work on generative inductive reasoning in the NLP domain, which is to generate natural language rules from specific natural language observations with language models, introducing the requirements on inductive reasoning from the philosophy of science community.
Motivated by the empirical experience that language models tend to generate vague and not specific rules, they additionally propose the fourth requirement:
(4) ``rule'' should be clear and in enough detail.
The fourth requirement might have been overlooked by the philosophy of science community since it's too obvious.
Motivated by the requirements, \citet{DBLP:journals/corr/abs-2212-10923} design an overly-generation-then-filtering mechanism, leveraging language models to first generate many preliminary rules and then filter those do not satisfy the requirements.
Then methods are developed to use self-refine to replace filtering and use more reasoning steps for better rules~\citep{DBLP:journals/corr/abs-2310-07064,DBLP:journals/corr/abs-2309-05660,DBLP:conf/nips/ZhongZLAKS23,DBLP:conf/iclr/QiuJLSPBWK0D024}.
%
%
However, the ``rules'' this line of works try to induce are either known knowledge, or not scientific knowledge but synthesized patterns.

\citet{DBLP:conf/acl/YangDLZPC24} make the first attempt to extend the classic inductive reasoning task setting~(to discover known/synthetic knowledge) into a real scientific discovery setting: to leverage LLMs to autonomously discover novel and valid social science scientific hypotheses from the publicly available web data. Specifically, they collect news, business reviews, and Wikipedia pages on social science concepts as the web data to discover hypothesis.

\citet{DBLP:conf/icml/MajumderS0HSC24,DBLP:journals/corr/abs-2407-01725} further propose the concept of ``data-driven discovery'', which is to discover hypotheses across disciplines with all the public experimental data on the web~(and private experimental data at hand).
Their motivation is that the potential of the large amount of publicly available experimental data has not been fully exploited that lots of novel scientific hypotheses could be discovered from the existing data.


\subsection{Development of Methods}
\label{sec:discovery_method}
Among the methods developed for scientific discovery, there is one clear method development trajectory.
We begin by introducing this trajectory, followed by an exploration of other methods.

\subsubsection{Main Trajectory}
\label{subsec:main_trajectory}

In general, this method development trajectory for scientific discovery can be seen as incorporating more key components into the methods.
Table~\ref{tab:discovery_method} summarizes the key components we identify as important and indicates whether each method incorporates them.
Specifically, they are ``strategy of inspiration retrieval'', ``novelty checker'', ``validity checker'', ``clarity checker'', ``evolutionary algorithm'', ``leverage of multiple inspiration'', ``ranking of hypothesis'', and ``automatic research question construction''.
Here, each ``key component'' refers to a detailed and unique methodology that has proven effective 
for scientific discovery tasks.
We exclude broad general concepts that may intuitively seem helpful but it's not clear how a specific method from the concept can be effective for this task~(e.g., tool usage).
Next, we introduce these key components.
For each key component, we use one or two paragraphs to give a short overview, summarizing its development trace.
The reference information for each method mentioned in this section can be found in Table~\ref{tab:discovery_method}.

\paragraph{Inspiration Retrieval Strategy}
In addition to relying on background knowledge, literature-based discovery (LBD) facilitates the retrieval of additional knowledge as a source of inspiration for formulating new hypotheses. 
SciMON~\citep{DBLP:conf/acl/0005DJH24} first introduces the concepts of LBD to the discovery task, demonstrating that new knowledge can be composed of linkage of existing knowledge. 
It is vital that the inspiration should not be known to be related to the background before, or at least should not be used to associate with the background in a known way~\citep{yang2024moose}. Otherwise, the hypothesis would not be novel.

Inspired by the ``ABC'' model in classic LBD formalization, given a background knowledge, SciMON retrieves semantically similar knowledge, knowledge graph neighbors, and citation graph neighbors as inspirations.
Specifically, two knowledge are identified as ``semantically similar'' if their embeddings from SentenceBERT~\citep{DBLP:conf/emnlp/ReimersG19} have high cosine similarity;
The knowledge graph they built follows a ``\textit{[method, used-for, task]}'' format.
ResearchAgent strictly follows the ``ABC'' model by constructing a concept graph, where a link represents the two connected concept nodes have appeared in the same paper before. 
It retrieves inspiration concepts that are connected with the background concepts on the concept graph~(concept co-occurence).
Scideator retrieves inspiration papers based on semantic matching~(semantic scholar API recommendations) and concept matching~(papers containing similar concepts in the same topic, same subarea, and different subarea).
SciPIP~\citep{Wang2024SciPIPAL} retrieves inspirations from semantically similar knowledge~(based on SentenceBERT), concept co-occurence, and citation graph neigbors.
It proposes filtering methods to filter not useful concepts for concept co-occurence retrieval.

Different from selecting semantic or citation neighbors as inspirations, SciAgents randomly sample another concept that is connected with the background concept in a citation graph~(via a long or short path) as the inspiration. 

MOOSE~\citep{DBLP:conf/acl/YangDLZPC24} proposes to use LLM-selected inspirations: given the research background and some inspiration candidates in the context, and ask an LLM to select inspirations for the research background from the candidates.
Then MOOSE-Chem~\citep{yang2024moose} also adopts it.
MOOSE-Chem assumes that after training on hundreds of millions of scientific papers, the most advanced LLMs might already have a certain level of ability to identify the inspiration knowledge for the background to compose a novel discovery of knowledge. 
MOOSE-Chem analyzes this assumption by annotating 51 chemistry papers published in 2024~(which are only available online in 2024) with their background, inspirations, and hypothesis, and see whether LLMs with training data up to 2023 can retrieve the annotated inspirations given only the background. 
Their results show a very high retrieval rate, indicating that the assumption could be largely correct.
Then Nova also adopts LLM-selected inspirations, with the motivation that leveraging the LLM’s internal knowledge to determine useful knowledge for new ideas should be able to surpass traditional entity or keyword-based retrieval methods.

\paragraph{Feedback Modules}
The next key component is the iterative feedback on the generated hypotheses in the aspects of novelty, validity, and clarity. 
These three feedbacks are first proposed by MOOSE, motivated by the requirements for a hypothesis in inductive reasoning~\citep{DBLP:journals/corr/abs-2212-10923,norton2003little}. 
These three aspects are objective enough to give feedback, and each of them is essential for a good hypothesis. 

\begin{table*}[t!]
\centering
\caption{Discovery Methods. Here ``NF'' = Novelty Feedback, ``VF'' = Validity Feedback, and ``CF'' = Clarity Feedback, ``EA'' = Evolutionary Algorithm, ``LMI'' = Leveraging Multiple Inspirations, ``R'' = Ranking, ``AQC'' = Automatic Research Question Construction. The order of methods reflect their first appearance time.}

\resizebox{\columnwidth}{!}{
\begin{tabular}{c|cccccccc}

\toprule
\textbf{Methods}     & \textbf{Inspiration Retrieval Strategy} & \textbf{NF} & \textbf{VF} & \textbf{CF} & \textbf{EA} &  \textbf{LMI} & \textbf{R} & \textbf{AQC} \\ \midrule
SciMON~\citep{DBLP:conf/acl/0005DJH24}      & Semantic \& Concept \& Citation Neighbors  & \checkmark              & -                 & -               & -                     & -                    & -   &  -   \\
MOOSE~\citep{DBLP:conf/acl/YangDLZPC24}       & LLM Selection                  & \checkmark              & \checkmark                & \checkmark              & -                     & -                    & -  &  \checkmark  \\
MCR~\citep{DBLP:conf/emnlp/SprueillEOSJC23}         & -                           & -          & \checkmark        & -   & -   & -  & \checkmark  & - \\ 
Qi~\citep{DBLP:journals/corr/abs-2311-05965}          & -                             & \checkmark              & \checkmark                & -               & -                     & -                    & -   &  - \\
FunSearch~\citep{DBLP:journals/nature/RomeraParedesBNBKDREWFKF24}   & -                             & -              & \checkmark                & -              & \checkmark                    & -                    & \checkmark  &  - \\
ChemReasoner~\citep{DBLP:conf/icml/SprueillEAOSJLJ24} & -                         & -        &  \checkmark    & -   & -   & -   & \checkmark  & - \\  
HypoGeniC~\citep{DBLP:journals/corr/abs-2404-04326}  &  -   &  -  &  \checkmark  &  -  &  -  &  -  &  \checkmark  &  -  \\
ResearchAgent~\citep{DBLP:journals/corr/abs-2404-07738}  &  Concept Co-occurrence Neighbors  &  \checkmark  &  \checkmark  & \checkmark  &  -  &   -  &  -  &  -  \\
LLM-SR~\citep{DBLP:journals/corr/abs-2404-18400}  &  -  &  -  &  \checkmark  &  -  &  \checkmark  &  -  &  \checkmark   &  -  \\
SGA~\citep{DBLP:conf/icml/MaWGSTRGM24} &  -           &      -    &   \checkmark    &  -    &  \checkmark & -   & -  &  - \\
AIScientist~\citep{lu2024ai} & -                             & \checkmark              & \checkmark                & -               & \checkmark                    & -                    & \checkmark  & \checkmark  \\
MLR-Copilot~\citep{li2024mlr}  &   -   &    -   &  -   &  -   &  -   &  -   &  -   &  \checkmark \\
IGA~\citep{DBLP:journals/corr/abs-2409-04109}  &   -    &   -  &  -   &   -   &  - &  -  &  \checkmark  &  -  \\
SciAgents~\citep{DBLP:journals/corr/abs-2409-05556}   & Random Selection               & \checkmark                & \checkmark                  & -                & -                     & -                    & -   & -  \\
Scideator~\citep{DBLP:journals/corr/abs-2409-14634}   &  Semantic \& Concept Matching    &   \checkmark   &  -    &   -     &   -    &   -    &   -  & - \\
MOOSE-Chem~\citep{yang2024moose}  & LLM selection                  & \checkmark              & \checkmark                & \checkmark              & \checkmark                    & \checkmark                   & \checkmark  & - \\
VirSci~\citep{su2024two}  &  -    &  \checkmark    &  \checkmark   &    \checkmark  &  -  &   -  &  -  &  \checkmark \\
CoI~\citep{li2024chain}    &    -    &  \checkmark & - & - & - & - & - & \checkmark \\
Nova~\citep{hu2024nova}    &   LLM selection  & -    &   -   &   -   &  -   &   \checkmark    &   -  & - \\
CycleResearcher~\citep{weng2024cycleresearcher}  &  -  &  -  &  -  &  -  &  - &  -  &  \checkmark   &   - \\
SciPIP~\citep{Wang2024SciPIPAL}  & Semantic \& Concept \& Citation Neighbors  &  -  &  -  & - & - &  -  &  -  &  - \\
\bottomrule
\end{tabular}}
\label{tab:discovery_method}
\end{table*}

\begin{itemize}

\item Novelty Checker.
The generated hypotheses should be a novel finding compared to the existing literature.
When a hypothesis tends to be similar to an existing hypothesis, feedback on enhancing its novelty could be beneficial for hypothesis formulation.
Existing methods for novelty feedback are all based on LLMs.
In general, there are three ways to provide novelty feedback.
The first method evaluates each generated hypothesis against a related survey (MOOSE); 
the second iteratively retrieves relevant papers for comparison (SciMON, SciAgents, Scideator, CoI); 
the third directly leverages the internal knowledge of LLMs for evaluation (Qi, ResearchAgent, 
AIScientist, MOOSE-Chem, VirSci).

\item Validity Checker.
The generated hypotheses should be valid science/engineering findings that precisely reflect the objective universe~\citep{norton2003little}.
A real validity feedback should be from the results of experiments.
However, it is time-consuming and costly to conduct experiments for each generated hypothesis.
Therefore, currently, validity feedback almost entirely relies on the heuristics of LLMs or other trained neural models.
The exceptions are FunSearch, HypoGeniC, LLM-SR, and SGA.
Specifically, FunSearch is about generating code for math problems. The compiler and verification code are naturally efficient and effective verifiers;
HypoGeniC and LLM-SR focus on data-driven discovery, which means they have access to observation examples that can be used to check consistency with each generated hypothesis;
SGA creates a virtual physical simulation environment to mimic real experiments. 
However, validity checker is still a significant challenge for the scientific discovery community. 
Future research directions include robotics and automation labs, which could automatically perform wet-lab experiments~(e.g., biology and chemistry experiments) to verify the generated hypotheses.
For computer science-related hypotheses, the future research direction could be more advanced systems for automatic code implementation.

\item Clarity Checker.
The generated hypotheses should be sufficiently clear in conveying information and provide adequate details~\citep{DBLP:journals/corr/abs-2212-10923}.
However, LLMs tend to generate hypotheses with 
insufficient details~\citep{DBLP:conf/acl/0005DJH24}.
Therefore, it would be beneficial to provide feedback in terms of clarity to refine the hypothesis and expand it with details~\citep{DBLP:conf/acl/YangDLZPC24}.
Current methods~(MOOSE, ResearchAgent, MOOSE-Chem, and VirSci) all adopt LLMs to provide self-assessment on clarity.

\end{itemize}

\paragraph{Evolutionary Algorithm}
Evolutionary Algorithm is a subset of optimization algorithms inspired by the principles of biological evolution.
It assumes the existence of an ``environment'', where an entity that can't adapt to it would be ``eliminated'', and super entity would be evolved from the ``recombination'' of characteristics between entities that have some adaptability to the environment~(this process is also called as ``mutation'').

This key component is important since 
(1) the real experiment evaluation and the heuristic evaluation of the generated hypotheses naturally serve as the ``environment''.
(2) the essence of scientific hypothesis discovery fundamentally can be seen as mutation to unknown yet valid knowledge from only known knowledge input. 
Although with similar goals, current scientific discovery methods leverage the evolutionary algorithm in different ways.

FunSearch first introduced the evolutionary algorithm to the scientific discovery task. 
They adopt an island-based evolutionary algorithm, where each island is a group of similar methods, and each island keeps mutating to new hypotheses.
At some time intervals, some least ranking islands are ``eliminated'', and new islands consisting of the best-performing hypotheses from every island are formed, encouraging the ``recombination'' between merits between islands.
LLM-SR adopts a similar island-based evolutionary algorithm.

SGA leverages it as ``evolutionary search'', which is to generate multiple offspring in each iteration and retain the best selection.
They also adopt an evolutionary crossover, where LLMs generate new hypotheses from various past experiments for better exploration.

MOOSE-Chem designs it as an ``evolutionary unit'', to better associate background knowledge and inspiration knowledge.
Specifically, given background and inspiration knowledge, they first generate multiple unique hypotheses to associate the two. Each hypothesis is then independently refined, and finally, the refined hypotheses are recombined to better integrate the background and inspiration knowledge into a cohesive hypothesis.
It encourages different mutation variants from the same input and gathers the advantages from each mutation variant.

\paragraph{Leveraging Multiple Inspirations}
Here the ``Leveraging Multiple Inspirations''~(LMI) component we discuss is about a clear identification of several inspirations, so that these identified inspirations will be all leveraged into the final hypothesis~(e.g., in a sequential way).
It is important, where different methods have different reasons.

MOOSE-Chem is the first to introduce this component, motivated by the observation that many 
disciplines such as chemistry and material science often require multiple inspirations to formulate a 
complete and publishable hypothesis. 
Specifically, they decompose the seemingly impossible-to-solve question $P(\text{hypothesis}|\text{research background})$ into many smaller, more practical and executable steps.
They do it by formulating a mathematical proof for the decomposition.
In general, the smaller steps involve identifying a starting inspiration, composing a preliminary hypothesis based on the background and inspiration, finding another inspiration to address gaps in the preliminary hypothesis, then composing an updated hypothesis with the new inspiration, and so on.
Their goal by utilizing multiple inspirations is to rediscover hypotheses in chemistry and material 
science that are published in high-impact journals such as Nature or Science.

In addition to MOOSE-Chem, Nova also retrieves multiple inspirations in a successive way, but with a different goal, which is to generate more diverse and novel research hypotheses.
Their motivation stems from IGA's experimental results that the diversity of generated hypotheses tends to saturate.
They identify one of the main reasons as that the input background information is the same, whereas incorporating different sets of inspirations can largely alleviate this issue by introducing flexible inputs.

\paragraph{Ranking of Hypotheses}
This key component is about providing a full ranking of the generated hypotheses.
It is important because LLMs can generate a large number of hypotheses in a short time, while real lab experiments to verify each of them are time-consuming and costly.
As a result, it would be very beneficial for scientists to know which hypothesis should be tested first.
Some methods~(e.g., MOOSE) adopt an automatic evaluation method to provide a preliminary understanding of generated hypotheses. 
The automatic evaluation method could naturally be used for ranking, but Table~\ref{tab:discovery_method} only focuses on how ranking is used in the methodology section~(but not in the automatic evaluation section). 

A majority of the methods adopt LLM's rated score as a reward value, which can be used for ranking~(MCR~\citep{DBLP:conf/emnlp/SprueillEOSJC23}, AIScientist, MOOSE-Chem, CycleResearcher). 
FunSearch focuses on a code generation problem, therefore can directly precisely evaluate the generated code by running them and checking results.
ChemReasoner~\citep{DBLP:conf/icml/SprueillEAOSJLJ24} finetunes a task-specific graph neural network model to obtain reward.
HypoGeniC~\citep{DBLP:journals/corr/abs-2404-04326} and LLM-SR~\citep{DBLP:journals/corr/abs-2404-18400} focuses on data-driven discovery, which means they have access to observation examples that can be used to check the consistency with the generated hypotheses, where the number of consistent examples can be used as the reward value for ranking.

Different from directly predicting a reward score, IGA takes a pairwise comparison, because they find that LLMs are poorly calibrated when asked directly to predict the final scores or decisions, but can achieve non-trivial accuracy when asked to judge which paper is better in pairwise comparisons.
Inspired by IGA~\citep{DBLP:journals/corr/abs-2409-04109}, CoI~\citep{li2024chain} proposes a pairwise automatic evaluation system, named Idea Arena. 
Nova~\citep{hu2024nova} also adopts a pairwise automatic evaluation method.

\paragraph{Automatic Research Question Construction} 
This key component is about automatic construction of research question, so that automated scientific discovery methods can use it as input to discover hypotheses.
It indicates a different role of LLM systems in scientific discovery: without it, an LLM serves as a copilot, relying on researchers to propose good research questions; with it, the system operates in a “full-self-driving” mode, capable of independent discovery without human input.
The “full-self-driving” mode was first introduced by MOOSE and framed as an “automated” setting for scientific discovery. 
Specifically, they adopt an LLM-based agent to continually search through the discipline-related web corpus to find interesting research questions.
AIScientist explores research directions by leveraging a starting code implementation as input. 
MLR-Copilot finds research directions by analyzing the research gaps from input papers.
SciAgents and Scideator skip research questions by directly generating hypotheses based on the pairing of concepts.
VirSci generates research questions by leveraging LLM-based scientist agents to brainstorm.
CoI finds research questions by collecting a development line of methods and then predicting the next step. 
Nova directly generates seed ideas from input papers and common idea proposal patterns, skipping the research question construction step.

\subsubsection{Other Methods}
\label{subsec:other_method}

In this section, we introduce the methods that are different from the methods in the ``main trajectory''~(\S~\ref{subsec:main_trajectory}).
These methods themselves are very diverse, focusing on different aspects of scientific discovery.
For example, \citet{DBLP:journals/corr/abs-2309-05689} leverage a distinct methodology, \citet{pu2024ideasynth} focus on HCI, \citet{liu2024literature} also consider the integration of experiment results, \citet{weng2024cycleresearcher,li2024learning} leverage reviews as preference learning to finetune the hypothesis proposer model.

\citet{DBLP:journals/corr/abs-2309-05689} try to use GPT-4 to tackle the very challenging research question: ``whether P = NP or not''.
They propose ``Socratic reasoning'', which encourages LLMs to recursively discover, solve, and integrate problems while facilitating self-evaluation and refinement.
Their method could be useful when trying to prove a very challenging existing hypothesis.

IdeaSynth~\citep{pu2024ideasynth} is a research idea development system, which represents idea concepts as linked nodes on a canvas.
Its effects are investigated in a human-computer interaction scenario.
They found through a lab study that human participants using IdeaSynth can explore more alternative ideas and expand initial ideas with more details compared to human participants using a strong LLM-based baseline.

\citet{liu2024literature} make the first attempt trying to unify literature-based discovery and data-driven discovery. 
Given an initial set of experiment results, it retrieves related literature and adopts an iterative refinement approach to keep improving a hypothesis to make it consistent with the experiment results and leverage findings from the retrieved literature.

\citet{weng2024cycleresearcher} propose a dual system that includes CycleResearcher and CycleReviewer, where the CycleResearcher is in charge of idea formulating and paper writing, and the CycleReviewer is in charge of scoring the written papers.
The dual system has a synergy that the scores from CycleReviewer can compose preference data to train CycleResearcher. 
The dual system only focuses on idea formulating and paper writing, skipping experiment planning and implementation.

\citet{li2024learning} propose fine-tuning LLMs to be better idea generators and introduce a novel framework that employs a two-stage approach combining Supervised Fine-Tuning (SFT) and Controllable Reinforcement Learning (RL). They focus on dimensions of feasibility, novelty and effectiveness. The dimensional controllers enable dynamic adjustment of the generation process.


\subsection{Benchmarks}
\label{sec:discovery_benchmark}

Overall the tasks in automated scientific discovery can be divided into ``literature-based discovery'' and ``data-driven discovery''. Researchers design different benchmarks for each task respectively.

\subsubsection{Literature-based Discovery}
Literature-based discovery is in general about connecting knowledge (pieces) in existing publications and associating them to create new knowledge. In this process, the knowledge to start with is from the research background.
A research background can be seen as consisting of two components:
(1) a research question, and 
(2) a background survey, which discusses the state-of-the-art methods or knowledge for the research question.
With the start knowledge in the research background, the other knowledge to connect is usually by searching through the existing publications.
Here the other knowledge is referred to as ``inspiration''~\citep{DBLP:conf/acl/0005DJH24,DBLP:conf/acl/YangDLZPC24}.
Then the research background and the retrieved inspiration(s) are associated to create a ``hypothesis''.

\begin{table*}[t!]
\caption{Discovery benchmarks aiming for novel scientific findings. The Biomedical data SciMON~\citep{DBLP:conf/acl/0005DJH24} collected is up to January 2024. 
RQ = Research Question; BS = Background Survey; I = Inspiration; H = Hypothesis.
\citet{DBLP:journals/corr/abs-2311-05965}'s dataset contains a train set where the publication date of the papers is before January 2023.
* in the date column represents the authors have checked the papers should not only be published after the date, but are also not available online before the date~(e.g., through arXiv).
The five disciplines \citet{kumar2024can} cover are Chemistry, Computer Science, Economics, Medical, and Physics.}
\label{tab:discovery_benchmark}
\resizebox{\columnwidth}{!}{
\begin{tabular}{c|cccccccc}
\toprule
\textbf{Name}        & \textbf{Annotator}    & \textbf{RQ} & \textbf{BS} & \textbf{I} & \textbf{H} & \textbf{Size}   & \textbf{Discipline}                    & \textbf{Date}              \\ \midrule
SciMON~\citep{DBLP:conf/acl/0005DJH24} & IE models    & \checkmark               & -                & -          & \checkmark        & 67,408 & NLP \& Biomedical             &      from 1952 to June 2022~(NLP)             \\
Tomato~\citep{DBLP:conf/acl/YangDLZPC24}      & PhD students & \checkmark               & -                & \checkmark         & \checkmark        & 50     & Social Science               & from January 2023 \\
\citet{DBLP:journals/corr/abs-2311-05965}          & ChatGPT      & -                & \checkmark               & -          & \checkmark        & 2900   & Biomedical                    & from August 2023~(test set)  \\
\citet{kumar2024can}       &  PhD students     &   -    &   \checkmark    &   -   &   \checkmark  & 100  & Five disciplines  & from January 2022 \\
Tomato-Chem~\citep{yang2024moose} & PhD students & \checkmark               & \checkmark               & \checkmark         & \checkmark        & 51     & Chemistry \& Material Science & from January 2024* \\
\bottomrule
\end{tabular}}

\end{table*}
Table~\ref{tab:discovery_benchmark} summarizes the literature-based discovery benchmarks, which aim for novel scientific findings.
The key components are the research question, background survey, inspiration identification, and hypothesis.
The hypotheses are collected from the ``abstract'' section~\citep{DBLP:conf/acl/0005DJH24}, the ``methodology'' section~\citep{DBLP:conf/acl/YangDLZPC24,yang2024moose}, or the ``future work'' and ``limitation'' sections~\citep{kumar2024can}.
Table~\ref{tab:discovery_benchmark} also includes the size of the dataset~(number of papers analyzed), disciplines of the papers, and the publication date of the papers.

The publication date is important to alleviate/avoid the data contamination problem.
The reason is that one of the main goals is to rediscover the groundtruth hypotheses, and the date can indicate which LLMs to use for the rediscovery~(its training data should be earlier than the date to avoid the potential data contamination problem).

Some of the benchmarks can be used for training since their large size~\citep{DBLP:conf/acl/0005DJH24,DBLP:journals/corr/abs-2311-05965}, while some are mainly used for evaluation since they are annotated by PhD students~\citep{DBLP:conf/acl/YangDLZPC24,kumar2024can,yang2024moose}.


\subsubsection{Data-driven Discovery}
\citet{DBLP:conf/icml/MajumderS0HSC24} propose the concept of ``data-driven discovery''. 
Here the ``data'' refers to the experiment results.
Their motivation is that given the ``observation'' of lots of (public and private) existing experimental results available online, LLMs might be able to find the general pattern of these data, where the general pattern could be a novel research hypothesis.
Given the relation between the specific observations and the general hypothesis, ``data-driven discovery'' is very related to the inductive reasoning task, where the observation space is the full publicly available experiment results on the web and the private experiment results at hand.

DiscoveryBench~\citep{DBLP:journals/corr/abs-2407-01725} is the first data-driven discovery benchmark. 
It comprises 264 discovery tasks extracted manually from over 20 published papers and 903 synthetic tasks. 
The input of the task consists of a research question and a set of experimental data.
The goal is to answer the research question with a hypothesis that can be supported by the experimental data.
It also introduces a structured formalism for the generated hypotheses, that the hypotheses should consist of three components: context, variables, and relationships.
Specifically, a hypothesis is about the relationships between the two variables under the context.

DiscoveryWorld~\citep{DBLP:journals/corr/abs-2406-06769} is the first discovery benchmark with a virtual environment.
The main motivation is twofold: (1) real-world experiments are costly and require substantial domain expertise; and (2) abstracting from task-specific details encourages the development of more general discovery methods. To address these challenges, it establishes a virtual environment for agents to discover hypotheses. 
It includes 120 different challenge tasks, where the hypotheses reflect the real patterns of the world.

\subsection{Evaluation Development Trend}
\label{sec:discovery_evaluation}

The evaluation methods for scientific discovery tasks are diverse. Arguably, nearly every paper proposes a new methodology uses a different evaluation approach. However, their metrics exhibit notable intersections, and some emerging trends in evaluation methods can be observed across these works.

The intersections of the \textbf{evaluation criteria} are ``novelty'', ``validity'', ``clarity'', and ``significance''. 
Some less-used evaluation criteria include ``relatedness'', ``interestingness'', and ``helpfulness''.
An alternative name for ``validity'' is ``feasibility''. They might be used interchangeably in many scenarios. 
``Validity'' refers to whether discovered scientific knowledge accurately reflects objective world, while ``feasibility'' concerns the practicability of an engineering finding. 
“Helpfulness” is a subjective evaluation, based on the idea that the goal of a discovery system is to act as a copilot for researchers; therefore, its perceived usefulness by researchers could be considered important. 
 
In terms of the \textbf{evaluator} selection, the evaluation methods can be divided into LLM-based and expert-based evaluation.
LLM’s direct evaluation has shown a high consistency score with expert evaluation in social science~\citep{DBLP:conf/acl/YangDLZPC24}.
However, in natural science disciplines such as chemistry, LLMs have been argued to lack the capability to provide reliable evaluations~\citep{DBLP:conf/icml/SprueillEAOSJLJ24}.
Expert evaluation is generally considered reliable. However, in challenging fields like chemistry, even 
an expert’s direct evaluation may lack sufficient reliability~\citep{yang2024moose}.
This is due to (1) the complexity of the discipline; and (2) the fact that slight changes in the research topic can necessitate entirely different background knowledge for evaluation, while experts typically have specialized research focuses, which may not cover the full range of knowledge required for a relatively reliable evaluation.

Based on the need for \textbf{reference}, evaluation methods can be categorized as direct evaluation and reference-based evaluation. 
Due to reliability concerns with direct evaluation, reference-based evaluation serves as an alternative~\citep{DBLP:journals/corr/abs-2407-01725,kumar2024can,yang2024moose}, which counts the key components from the ground truth hypotheses mentioned in the generated hypotheses.

Moreover, in addition to directly assigning a scalar evaluation score to a generated hypothesis, \citet{DBLP:journals/corr/abs-2409-04109} propose comparison-based evaluations to alleviate the incapacity of LLM-based evaluation of direct scoring: the LLM evaluator is asked to keep comparing pairs of generated hypotheses until a ranking is possible.
It can be used when comparing the quality of hypotheses generated by two methods, but might not help in judging the absolute quality of a hypothesis.

However, the ultimate evaluation should be only through real~(wet-lab) experiments. 
It raises challenges in the robotics and automatic experiment implementation fields.

\subsection{Major Progresses/Achievements on Discovering Hypotheses}


\citet{DBLP:conf/acl/YangDLZPC24} are the first to demonstrate that LLMs are capable of generating novel and valid scientific hypotheses, as confirmed through expert evaluation.
They find three social science PhD students to directly evaluate the novelty and validity of the generated social science hypotheses.
Then \citet{DBLP:journals/corr/abs-2409-04109} provide the first large-scale expert evaluation on LLM-generated hypotheses by hiring 100+ NLP researchers.
They find a statistically significant conclusion that LLM can generate more novel but slightly less valid research hypotheses than human researchers.
Then \citet{yang2024moose} show that an LLM-based framework can rediscover the main innovations of many chemistry and material science hypotheses published in Nature, Science, or a similar level in 2024~(the hypotheses are only available online in 2024), using LLMs trained exclusively on data available until October 2023.

\subsection{Challenges and Future Work}

\paragraph{Challenge}
Scientific discovery is to find novel knowledge that has not been verified by wet lab experiments.
In some disciplines such as chemistry, even an expert's evaluation of the generated novel hypothesis is not reliable enough. 
This causes a need for automated experiments conduction to verify the large-scale machine-generated hypotheses.

In addition, current methods on scientific discovery highly rely on the ability of existing available LLMs.
LLMs with better capacity on the universal tasks usually can also lead to discovered hypotheses with better quality~\citep{DBLP:conf/acl/YangDLZPC24}.
As a result, LLM-based discovery methods may have an upper performance limit, constrained by the capabilities of the state-of-the-art LLMs.
However, it is largely~(if not completely) unclear how should we augment LLM's ability on the task of scientific discovery.

Thirdly, it is unclear on a sufficient set of internal reasoning structure for scientific discovery: current works rely heavily on retrieving from high-quality knowledge source~(e.g., literature) as inspiration to generate hypothesis. 
But it is unclear on whether there are any more internal reasoning structures that can help with the process.

Finally, building accurate and well-structured benchmark highly relies on experts. 
However, the size of a expert-composed benchmark is usually very limited. 
It is unclear on how should we scale up an accurate and well-structured discovery-oriented benchmark.

\paragraph{Future work}
The first line of future work is to enhance automated experimental execution, as it remains the most reliable way to test the validity of a hypothesis. 
This process may vary across disciplines. 
In computer science, the bottleneck might be the coding ability, especially the ability to program a large system.
In chemistry or biology, the bottleneck might lie in the robotics methods to conduct experiments~\citep{boiko2023autonomous}.

The second line of future work is to enhance the LLM's ability in hypothesis generation. 
Currently, it is still not very clear how to increase this ability.
The aspects might include training data collection methods and training strategies.

The third line of future work is to investigate other internal reasoning structures of the scientific discovery process.
This might need an interdisciplinary effort, involving the philosophy of science~(also known as science of science)~\citep{fortunato2018science}.

The fourth line of future work is to investigate how to leverage LLMs to automatically collect accurate and well-structured benchmark.

\section{LLMs for Experiment Planning and Implementation}
\label{section4}

\subsection{Overview}

Beyond generating hypotheses, LLMs are increasingly employed in scientific research to automate experiment design and streamline workflows. LLMs poss comprehensive internal world knowledge, enabling them to perform informed actions in real world without training on specific domain data. To maximize their potential, LLMs are designed in an agent-based fashion with two critical properties~\citep{KambhampatiVGVS24}: modularity and tool integration. Modularity ensures that LLMs can interact seamlessly with external systems, such as databases, experimental platforms, and computational tools, while tool-augmented frameworks enable LLMs to serve as central controllers in workflows interfaced with specialized modules for data retrieval, computation, and experimental control. 
This section explores how LLMs are applied specifically to support the \textbf{planning} and \textbf{implementation} of research ideas.

\subsection{Optimizing Experimental Design}
\label{section4.2}
LLMs are transforming the experimental design process by enabling more efficient and adaptive workflows in scientific research. Their capacity to process and analyze extensive datasets empowers researchers to decompose complex tasks, select optimal methodologies, and enhance the overall structure of experiments. This section explores how LLMs facilitate experimental design optimization across various domains.

Task decomposition involves breaking experiments into smaller, manageable sub-tasks, a process often necessitated by the complexity of real-world research to ensure alignment with specific research goals \citep{huang2024understanding}. 
Numerous studies \citep{rasal2024navigating, boiko2023autonomous, shen2024hugginggpt, m2024augmenting, autogen2023, CRISPR-GPT} demonstrate how LLMs simplify intricate problems by defining experimental conditions and specifying desired outputs. For instance, HuggingGPT \citep{shen2024hugginggpt} utilizes LLMs to parse user queries into structured task lists while determining execution sequences and resource dependencies. Similarly, CRISPR-GPT \citep{CRISPR-GPT} automates CRISPR-based gene-editing experiment design by facilitating the selection of appropriate CRISPR systems, designing guide RNAs, recommending cellular delivery methods, drafting protocols, and planning validation experiments. ChemCrow \citep{m2024augmenting} employs iterative reasoning and dynamic planning, using a structured "Thought, Action, Action Input, Observation" loop \citep{yaoreact} to refine its approach based on real-time feedback. Multi-LLM systems, such as Coscientist \citep{boiko2023autonomous} and LLM-RDF \citep{ruan2024automatic}, further leverage specialized agents to extract methodologies from literature, translate natural language descriptions into standardized protocols, generate execution code for automated platforms, and adaptively correct errors during execution.

Advanced prompting-based techniques such as in-context learning, Chain of Thought~\cite{weichain} and ReAct~\citep{yaoreact}, are often employed in studies described above  to enhance the reliability and accuracy of experimental planning in LLM-assisted workflows. Moreover, LLMs are also capable of enhancing experimental design through reflection and refinement~\citep{madaan2024self, shinn2024reflexion}, a process that allows them to continuously evaluate and improve experimental plans. 
For instance, by simulating expert discussions, LLMs engage in a collaborative dialogue \cite{liiqa}, challenging assumptions, and refining their output through iterative analysis~\citep{li2024simulating}. This method mirrors real-world scientific problem solving, where discrepancies between expert opinions foster a deeper exploration of the problem space, and consensus is achieved through rigorous debate and synthesis of diverse perspectives.

\subsection{Automating Experimental Processes}
\label{section4.3}

LLMs revolutionize scientific research by automating repetitive and time-consuming tasks in experimental processes. This automation significantly enhances productivity, allowing researchers to delegate labor-intensive processes such as data preparation, experimental execution, analysis, and reporting to LLM-based systems~\citep{wang2024survey}.

\subsubsection{Data Preparation}
\label{section4.3.1}
One of the most labor-intensive aspects of research is data preparation, which includes tasks such as cleaning~\citep{zhang2023jellyfish, chen2024data}, labeling~\citep{tan2024large, ziemsetal2024large}, and feature engineering~\citep{hollmann2024large}. Large Language Models (LLMs) can automate these processes, especially when dealing with large datasets where manual data curation would be inefficient. Additionally, in situations where data is difficult to obtain, LLMs can synthesize experimental data directly~\citep{Li2023AreYI, liu2023training, Li2024FinegrainedHD}. For instance, in social science, where conducting experiments with human subjects is often expensive and unethical, \citet{liu2023training} design a sandbox to simulate a social environment and deployed multiple agents (LLMs) to interact with each other. This approach allows researchers to collect data on agent social interactions for subsequent analysis.

\subsubsection{Experiment Execution and Workflow Automation}
\label{section4.3.2}
 


To automate the experimental workflow in scientific research, LLM-based agents can acquire task-specific capabilities through a combination of 
pretraining~\citep{RivesMSGLLGOZMF21,lin2023evolutionary}, fine-tuning~\citep{FerruzH22,he2024novo}, and tool-augmented learning. Pretraining on extensive dataset provides foundational knowledge, while fine-tuning on domain-specific datasets refines this knowledge for targeted scientific applications. To enhance task execution, LLMs are often coupled with domain-specific knowledge bases~\citep{m2024augmenting,boiko2023autonomous,Wang2024Efficient} or preconfigured workflows~\citep{Liu2024,boiko2023autonomous}. Advanced prompting techniques like in-context learning and chain-of-thought prompting~\citep{DrugAssist,Liu2024} enable LLMs to quickly adapt to new experimental protocols.
Additionally, iterative adjustments with task-specific feedback loops allow the LLM to refine its outputs based on experimental goals~\citep{Ramos2023,DrugAssist}.

Based on these principles, LLM plays a diverse role in automating experimental workflows across different disciplines. 
In chemistry, ChemCrow~\citep{m2024augmenting}, an LLM chemistry agent, leverages 18 expert-designed tools to autonomously plan and execute complex chemical syntheses, bridging computational and experimental domains. Similarly, Coscientist~\citep{boiko2023autonomous} integrates LLM with lab automation to optimize reactions like palladium-catalyzed syntheses. LLMs have also been employed for evolutionary search strategies to explore vast chemical spaces~\citep{Wang2024Efficient}, enabling the identification of candidate molecules while reducing experimental burdens. \citet{Ramos2023} combine natural language inputs with Bayesian optimization for catalyst synthesis, streamlining iterative design cycles. Furthermore, LLMs have been utilized for hypothetical scenario testing and reaction design, minimizing experimental iterations through hypothesis pre-screening~\citep{DBLP:conf/emnlp/SprueillEOSJC23, DBLP:conf/icml/SprueillEAOSJLJ24}.
In drug discovery, ChatDrug~\citep{Liu2024} integrates modules for prompting, retrieval, and domain feedback to facilitate drug editing, while DrugAssist~\citep{DrugAssist} iteratively optimizes molecular structures through human-machine dialogue.
In biological and medical research, Models like ESM-1b~\citep{RivesMSGLLGOZMF21} and ESM-2~\citep{lin2023evolutionary} encode protein sequences, capturing structural properties for predictive tasks, such as secondary and tertiary structure predictions, eliminating the need for labor-intensive experiments. By fine-tuning LLMs on protein families, \citet{FerruzH22} generate highly divergent yet functional protein sequences. Additionally, \citet{he2024novo} introduce an antibody generative LLM for de novo SARS-CoV-2 antibody design, achieving specificity and diversity while reducing reliance on natural antibodies.

\subsubsection{Data Analysis and Interpretation}
\label{section4.3.3}

Beyond automating the execution of experiments, LLMs assist in data analysis by generating natural language explanations and constructing meaningful visualizations, which are essential for interpreting complex datasets and ensuring that the derived insights are accessible and actionable~\citep{singh2024rethinking}.
Traditionally, data analysis requires extensive statistical expertise, manual calculation, and the interpretation of large volumes of experimental results. LLMs simplify this by automating tasks such as statistical modeling and hypothesis testing. For instance, ~\citet{li2024automated} demonstrate that LLMs can serve as modelers, proposing, fitting, and refining probabilistic models based on real-world data, while also providing critical feedback on model performance through techniques like posterior predictive checks.
Additionally, LLMs excel at uncovering hidden patterns, trends, and relationships within textual data. In social media data analysis, LLMs provide insights into public sentiment and emerging trends~\citep{yang2024mentallama}, and in environmental data interpretation, they contribute to improved understanding and decision-making in environmental science~\citep{oliver2024opening}.
Moreover, they are also instrumental in thematic analysis~\citep{dai2023llm, rasheed2024can}, helping to identify themes and patterns in qualitative data. Their application extends to financial data analysis as well, where they enhance forecasting and risk assessment capabilities~\citep{zhao2024revolutionizing}. AutoGen~\citep{autogen2023} provides a generic framework that enables the creation of diverse applications using multiple customizable agents (LLMs). These agents can interact through natural language and code, supporting a wide range of downstream task such as data modeling and data analysis~\citep{jing2024dsbenchfardatascience}.

\subsection{Benchmarks}
\label{section4.4}

\begin{table*}[!ht]
\centering
\caption{Benchmark for LLM-Assisted Experiment Planning and Implementation. ED = Optimizing Experimental Design, DP = Data Preparation, EW = Experiment Execution \& Workflow Automation, DA = Data Analysis \& Interpretation. ``General" in discipline means a benchmark is not designed for a particular discipline.}
\label{tab:llm_experiment_benchmarks}
\resizebox{\textwidth}{!}{
\begin{tabular}{>{\centering\arraybackslash}m{0.25\textwidth}|>{\centering\arraybackslash}m{0.03\textwidth}>{\centering\arraybackslash}m{0.03\textwidth}>{\centering\arraybackslash}m{0.03\textwidth}>{\centering\arraybackslash}m{0.03\textwidth}>{\centering\arraybackslash}m{0.20\textwidth}>{\centering\arraybackslash}m{0.50\textwidth}}
\toprule
\rowcolor{white}
\textbf{Benchmark Name} & \textbf{ED} & \textbf{DP} & \textbf{EW} & \textbf{DA} & \textbf{Discipline} & \textbf{Additional Task Details}  \\
\hline
TaskBench~\citep{shen2023TaskBench}&  \checkmark  &  - & - & - & General & Task decomposition, tool use  \\
DiscoveryWorld~\citep{DBLP:journals/corr/abs-2406-06769} & $\checkmark$ & - & $\checkmark$ & $\checkmark$ & General & Hypothesis generation, design \& testing \\
MLAgentBench \citep{huang2024mlagentbenchevaluatinglanguageagents} & $\checkmark$ & $\checkmark$ & $\checkmark$ & - & Machine Learning & Task decomposition, plan selection, optimization \\
AgentBench \citep{liu2023agentbenchevaluatingllmsagents} & $\checkmark$ & - & $\checkmark$ & $\checkmark$ & General & Workflow automation, adaptive execution \\
Spider2-V \citep{cao2024spider2vfarmultimodalagents} & - & - & $\checkmark$ & - & Data Science \& Engineering & Multi-step processes, code \& GUI interaction \\
DSBench \citep{jing2024dsbenchfardatascience} & - & $\checkmark$ & - & $\checkmark$ & Data Science & Data manipulation, data modeling\\
DS-1000 \citep{lai2022ds1000naturalreliablebenchmark} & - & $\checkmark$ & - & $\checkmark$ & Data Science & Code generation for data cleaning \& analysis \\
CORE-Bench \citep{siegel2024corebenchfosteringcredibilitypublished} & - & - & - & $\checkmark$ & Computer Science, Social Science \& Medicine & Reproducibility testing, setup verification \\
SUPER \citep{bogin2024superevaluatingagentssetting} & - & $\checkmark$ & $\checkmark$ & - & General & Experiment setup, dependency management \\
MLE-Bench \citep{chan2024mlebenchevaluatingmachinelearning} & - & $\checkmark$ & $\checkmark$ & $\checkmark$ & Machine Learning & End-to-end ML pipeline, training \& tuning \\
LAB-Bench~\citep{Laurent2024LABBenchMC}&  -  &  - &  \checkmark & \checkmark & Biology & Manipulation of DNA and protein sequences \\
ScienceAgentBench~\citep{chen2024scienceagentbench}&  -  &  \checkmark &  \checkmark & \checkmark & Data Science & Data visualization, model development \\
 \bottomrule
\end{tabular}}

\end{table*}

Benchmarks are essential for evaluating how effectively LLMs can support various aspects of experimental workflows. While not specifically created for LLM-assisted experiment implementation, many benchmarks are versatile enough to be applied to these tasks. For example, MLAgentBench~\citep{huang2024mlagentbenchevaluatinglanguageagents} covers task decomposition by helping break down complex research tasks, data handling by automating processes like data loading and transformation, and workflow management by optimizing machine learning experiment execution.

These benchmarks provide different venues and thus vary in their approaches. Evaluation methods range from task success rate, accuracy and execution consistency to comparisons with human benchmarks. These differences highlight the diverse ways LLMs can be integrated into research processes. Further details are presented in Table \ref{tab:llm_experiment_benchmarks}.

\subsection{Challenges and Future Work}
\paragraph{Challenges} 

The challenges of employing LLMs for experiment planning and implementation arise both from their intrinsic limitations and their application to domain-specific tasks. One fundamental limitation is their planning capability. As clarified by \citet{KambhampatiVGVS24}, LLMs in autonomous modes often fail to generate executable plans. They are prone to hallucinations, which can lead to irrational plans, deviations from task prompts, or an inability to follow complex instructions~\citep{huang2024understanding}. 
Prompt robustness poses another critical challenge in multi-stage experimental contexts. Minor variations in prompt wording, even when conveying the same intent, can result in inconsistent guidance throughout the planning and execution process~\citep{DBLP:conf/eacl/ZhuoLHSWHL23}, potentially affecting experimental outcomes. Additionally, the slow processing speed of autoregressive LLMs can impede real-time feedback in iterative and multi-step experiment planning, limiting their efficiency.
Application-specific challenges include difficulties in adapting to specialized roles, as LLMs struggle to emulate domain-specific scientific expertise and cognitive processes essential for generalizability across research domains~\citep{weng2024cycleresearcher}. For example, certain experiments may require simulating ethically sensitive or error-prone scenarios, which often conflict with the safety-aligned values embedded in LLMs.

\paragraph{Future work} 

Future research should address these challenges by enhancing core model capabilities and tailoring them to the unique requirements of experimental tasks. To mitigate hallucination risks, robust verification mechanisms can be integrated into workflows, such as cross-referencing outputs with external sound verifiers~\citep{KambhampatiVGVS24} or employing real-time feedback loops to correct inaccuracies dynamically~\citep{DBLP:journals/csur/JiLFYSXIBMF23}.
Improving prompt robustness may involve developing adaptive systems that monitor and modify prompt structures in response to contextual changes, ensuring consistency across planning stages. Efficiency enhancements could be achieved by creating faster, distilled versions of LLMs optimized for multi-step reasoning or hybrid systems combining LLMs with smaller, task-specific models to balance speed and accuracy.
For more effective role adaptation, fine-tuning LLMs with high-quality domain-specific datasets or developing modular frameworks could enable more precise emulation of specialized scientific reasoning. Additionally, designing adaptive alignment protocols may allow LLMs to safely simulate ethically complex scenarios  when addressing specific experimental goals.

\section{LLMs for Scientific Paper Writing}
\label{section5}
\subsection{Overview}
This section explores the integration of LLMs in three key areas of scientific paper writing: citation text generation (\S~\ref{section5.2}), related work generation (\S~\ref{section5.3}), and drafting and writing (\S~\ref{section5.4}). We examine the methodologies used, the effectiveness of these models, and the challenges faced in automating scientific writing. In addition, we discuss the evaluation metrics and benchmarks used in these tasks.

\subsection{Citation Text Generation}
\label{section5.2}
Given the context of a citing paper, citation text generation task aims to produce accurate textual summaries for a set of papers-to-cite.
LLMs have been pivotal in automating various aspects of citation text generation by providing rich contextual understanding and coherence, employing a range of methodologies to enhance both accuracy and usability. 
A pilot study by~\citet{xing-etal-2020-automatic} uses a pointer-generator network that can copy words from the manuscript and the abstract of the cited paper based on cross-attention mechanisms to generate citation texts. \citet{li2024explaining} prompt an LLM to generate a natural language description that emphasized the relationships between pairs of papers in the citation network.
On the other hand, models like AutoCite~\citep{Wang2023AutoCite} and BACO~\citep{Ge2021BACOAB} extend this work by adopting a multimodal approach, combining citation network structures with textual context to produce contextually relevant and semantically rich citation texts.
Furthermore, \citet{gu-hahnloser-2024-controllable, math10101763} allow users to specify attributes such as citation intent and keywords, integrating these into a structured template and fine-tuning an LM to generate citation texts that align with their needs.

\subsection{Related Work Generation}
\label{section5.3}
This task involves creating a related work section for a scientific paper based on cutting-edge reference papers~\citep{hoang-kan-2010-towards}. Compared to traditional multi-document summarization models~\citep{hu-wan-2014-automatic, chen2021capturing}, LLMs have demonstrated remarkable capabilities in handling the extensive input lengths characteristic of scientific documents and providing a rich contextual understanding. The success of LLMs in various natural language understanding and generation tasks, combined with their large context windows, has recently enabled more comprehensive and nuanced literature reviews, facilitating deeper insights and connections across diverse research areas.

\citet{Zimmermann2024Leveraging, martin2024shallow} develop case studies to explore the use of ChatGPT for literature review tasks and related work generation, showcasing its ability to assist researchers by quickly scanning large datasets of scientific publications and generating initial drafts of related work sections. However, directly employing LLMs in academic writing could lead to issues such as hallucinations, where the generated content is not grounded in factual data and may fail to accurately reflect state-of-the-art research.
To address these issues, numerous works have operated on the principle of Retrieval-Augmented Generation (RAG)~\citep{Lewis2020RAG}, which enhances LLM-based literature review generation by grounding in factual content retrieved from external sources~\citep{shi2023towards, Agarwal2024LitLLMAT,hu2024hireviewhierarchicaltaxonomydrivenautomatic,Yu2024Reinforced,susnjak2024automating}.
For instance, LitLLM~\citep{Agarwal2024LitLLMAT} utilize RAG to retrieve relevant papers on websites and re-rank them, reducing the time and effort needed for comprehensive literature reviews while minimizing hallucinations. 
HiReview~\citep{hu2024hireviewhierarchicaltaxonomydrivenautomatic} takes this further by integrating RAG-based LLMs with graph-based hierarchical clustering. This system first retrieved relevant sub-communities within a citation network and generated a hierarchical taxonomy tree. LLMs then generate summaries for each cluster, ensuring complete coverage and logical organization. 
\citet{nishimura-etal-2024-toward} integrate LLMs to emphasize novelty statement in related work sections. By comparing the new research with existing works, the LLMs help generate related work sections that explicitly highlight what is new and different, contributing to a more impactful comparison between the target paper and prior literature.

\subsection{Drafting and Writing}
\label{section5.4}

In the field of automated scientific writing, LLMs are being used across various tasks, ranging from generating specific textual elements to producing entire research papers. For more specific writing tasks, \citet{august-etal-2022-generating} propose to generate scientific definitions with controllable complexity tailored to different audiences, while SCICAP~\citep{hsu2021scicap} automates the generation of captions for scientific figures, enabling quick and accurate descriptions of visual data.
More holistic systems, such as PaperRobot~\citep{DBLP:conf/acl/WangHJKJBL19}, introduce an incremental drafting approach, where LLMs help organize and draft sections of a paper based on user inputs. Similarly, CoAuthor~\citep{Lee2022CoAuthor} takes a collaborative human-AI approach, in which LLMs help authors by generating suggestions and expanding text.
For fully autonomous writing, \citet{ifargan2024autonomous} explore how LLMs can generate complete research papers from data analysis to final drafts, while AutoSurvey~\citep{wang2024autosurvey} demonstrates the ability of LLMs to autonomously write comprehensive surveys by synthesizing and organizing existing research. Lastly, AI Scientist~\citep{lu2024ai} and CycleResearcher~\citep{weng2024cycleresearcher} propose an even broader system that not only drafts scientific papers but also contributes to the entire scientific process, including hypothesis generation and experiment design, highlighting the potential for fully automated scientific discovery and writing. 

\subsection{Benchmarks}
\label{section5.5}
We summarize the evaluation methods of automated scientific paper writing systems in three key fields: citation text generation, related work generation, and drafting and writing. In Table~\ref{tab:paper_writing_evaluation}, we provide a comprehensive summary of the specific datasets, metrics, and benchmarks for each task.

\begin{table*}[h!]
\centering
\caption{Evaluation Methods for automated paper writing, which includes three subtasks: citation text generation, related work generation, and drafting and writing. For the related work generation, there is no universally recognized benchmark.}
\label{tab:paper_writing_evaluation}
\renewcommand{\arraystretch}{1.2} 
\resizebox{\textwidth}{!}{
\begin{tabular}{m{0.12\textwidth} | m{0.2\textwidth} m{0.4\textwidth} m{0.6\textwidth}}

\toprule
\rowcolor{white}
\textbf{Task} & \textbf{Benchmark} & \textbf{Dataset} & \textbf{Metric} \\

\hline

\multirow{3}{0.1\textwidth}{Citation Text Generation} &
    ALEC~\citep{gao-etal-2023-enabling} & 
    ASQA~\citep{stelmakh-etal-2022-asqa}, QAMPARI~\citep{amouyal-etal-2023-qampari}, ELI5~\citep{fan-etal-2019-eli5} & 
    Fluency: MAUVE~\citep{pillutla2021mauve}, Correctness: precision, recall. Citation quality: citation recall, citation precision~\citep{gao-etal-2023-enabling} \\
\cline{2-4}

&   
    CiteBench~\citep{Funkquist2022CiteBenchAB} & 
    \citet{abura2020automatic}, \citet{chen2021capturing}, \citet{lu-etal-2020-multi-xscience}, \citet{xing-etal-2020-automatic} & 
    Quantitative: ROUGE~\citep{lin-2004-rouge}, BertScore~\citep{Zhang2019BERTScoreET}, Qualitative: citation intent labeling~\citep{cohan-etal-2019-structural}, CORWA tagging~\citep{li-etal-2022-corwa} \\

\midrule

Related Work Generation & \multicolumn{1}{l}{None}   & 
AAN~\citep{radev-etal-2009-acl}, SciSummNet~\citep{Yasunaga2019ScisummNet}, Delve~\citep{Akujuobi2017DelveAD}, S2ORC~\cite{lo-etal-2020-s2orc}, CORWA~\citep{li-etal-2022-corwa} & 
ROUGE~\citep{lin-2004-rouge}, BLEU~\citep{Papineni2002BleuAM}, Human evaluation: fluency, readability, coherence, relevance, informativeness \\

\midrule

\multirow{3}{0.1\textwidth}{Drafting and Writing} & 
    SciGen~\citep{moosavi2021scigen} & SciGen~\citep{moosavi2021scigen} & 
    BLEU~\citep{Papineni2002BleuAM}, METEOR~\citep{banerjee-lavie-2005-meteor}, MoverScore~\citep{Zhao2019MoverScoreTG}, BertScore~\citep{Zhang2019BERTScoreET}, BLEURT~\citep{sellam-etal-2020-bleurt}, Human evaluation: recall, precision, correctness, hallucination \\

\cline{2-4}

& SciXGen~\citep{chen-etal-2021-scixgen-scientific} & SciXGen~\citep{chen-etal-2021-scixgen-scientific} & 
BLEU~\citep{Papineni2002BleuAM}, METEOR~\citep{banerjee-lavie-2005-meteor}, MoverScore~\citep{Zhao2019MoverScoreTG}, Human evaluation: fluency, faithfulness, entailment and overall \\
\bottomrule

\end{tabular}}

\end{table*}

\paragraph{Citation Text Generation} The ALCE~\citep{gao-etal-2023-enabling} benchmark is the primary standard. Assessment of systems on three dimensions: fluency, correctness, and quality of citation text. ALCE is designed to test the ability of models to generate long-form answers with accurate citations across diverse domains. Their datasets cover a wide range of question types, with corpora spanning from Wikipedia to web-scale document collections. CiteBench~\citep{Funkquist2022CiteBenchAB}  is another benchmark that unifies multiple existing tasks to standardize the evaluation of citation text generation across various designs and domains, using both qualitative and quantitative metrics.

\paragraph{Related Work Generation} Currently, no single benchmark is universally recognized for this task, due to the vast differences in task definitions and simplifying assumptions in various studies~\citep{Li2024RelatedWA}. However, most works are built on corpus-level datasets, and commonly used sources of scientific articles include: ACL Anthology Network (AAN) Corpus~\citep{radev-etal-2009-acl}, SciSummNet~\citep{Yasunaga2019ScisummNet}, Delve~\citep{Akujuobi2017DelveAD}, Semantic Scholar Open Research Corpus (S2ORC)~\citep{lo-etal-2020-s2orc} and Citation Oriented Related Work Annotation (CORWA)~\citep{li-etal-2022-corwa}.
The summarization metric ROUGE~\citep{lin-2004-rouge} is the most frequently employed for automatic evaluation, with some work also using the translation metric BLEU~\citep{Papineni2002BleuAM}. Furthermore, human evaluations often rate fluency, readability, coherence with the target paper, and relevance and informativeness to the cited work on a five-point Likert scale.

\paragraph{Drafting and Wrigting} SciGen~\citep{moosavi2021scigen} benchmark supports the evaluation of reasoning-aware text generation from scientific tables, highlighting the challenges of arithmetic reasoning in text generation.
SciXGen~\citep{chen-etal-2021-scixgen-scientific}, another key benchmark, evaluates the context-aware text generation, focusing on the integration of external information into the generated text. 
Both SciGen and SciXGe use metrics like BLUE~\citep{Papineni2002BleuAM}, METEOR~\citep{banerjee-lavie-2005-meteor} and MoverScore~\citep{Zhao2019MoverScoreTG}, along with human evaluation.


\subsection{Challenges and Future Work}

\paragraph{Challenges}
The challenges in citation text generation, related work generation, and drafting and writing primarily arise from inherent limitations in LLMs, such as maintaining factual accuracy, ensuring contextual coherence, and handling complex information. LLMs often struggle with hallucinations~\citep{DBLP:journals/csur/JiLFYSXIBMF23}, generating incorrect or irrelevant citations, and are constrained by the retrieval systems they depend on~\citep{huang2023survey}. Limited context windows further restrict models' ability to manage extensive references or integrate relevant literature comprehensively~\citep{wang2024autosurvey}, potentially leading to incorrect citation ordering and inappropriate citation grouping. Additionally, ensuring scientific rigor and avoiding reliance on superficial or trivial sources remain persistent obstacles, as LLMs struggle to capture the depth and reasoning needed in academic writing~\citep{lu2024ai}.

Furthermore, the use of LLMs in academic writing introduces significant ethical concerns, particularly regarding academic integrity and plagiarism~\citep{Li2024RelatedWA}. This blurs the lines of authorship, as researchers might present machine-generated text as their own work. LLMs can also generate text that closely mimics existing literature, raising the risk of unintentional plagiarism where the generated content may not be sufficiently original. The convenience of using LLMs to draft sections of papers can undermine the rigorous intellectual effort traditionally required in academic writing, potentially devaluing the learning process and critical thinking skills essential to scholarly research.

\paragraph{Future Work} 
To overcome these challenges, future advancements should focus on improving retrieval systems and enhancing models' capacity to synthesize information from diverse, long-context sources~\citep{ Li2022AutomaticRW}. This includes developing better citation validation mechanisms, improving multi-document synthesis, and introducing real-time literature discovery to keep generated content up to date.
Additionally, incorporating domain-specific fine-tuning and reasoning-aware models will help generate more accurate, contextually relevant scientific text~\citep{moosavi2021scigen}. Fine-grained control over the writing process, such as adjusting tone and style, will also be crucial for improving the adaptability of LLMs to different academic needs~\citep{chen-etal-2021-scixgen-scientific, gao-etal-2023-enabling, lu2024ai}.
Furthermore, integrating human-in-the-loop systems, where human oversight and intervention are essential parts of the writing process, can ensure that the intellectual rigor and critical thinking inherent in scholarly work are preserved~\citep{Li2024RelatedWA, martin2024shallow}. Finally, to address the potential ethical concerns, it is crucial for the academic community to establish clear guidelines and ethical standards for the use of LLMs to ensure the integrity and originality of academic work.

\section{LLMs for Peer Reviewing}
\label{section6}

\subsection{Overview}
\label{section6.1}
Peer review is the cornerstone of scientific research. The integration of LLMs into the peer review process represents a significant advancement, addressing longstanding challenges such as reviewer bias, inconsistent standards, and workload imbalances \cite{Goldberg_2023_PeerReviewsOfPeerReviews, Piniewski_2024_EmergingPlagiarismPeerReview}. This integration has gained significant traction in the academic community, as evidenced by major computer science conferences adopting LLM-assisted reviewing practices. For instance, ICLR 2025 has announced the implementation of LLM-based systems to support reviewers in their evaluation process\footnote{\url{https://blog.iclr.cc/2024/10/09/iclr2025-assisting-reviewers/}}.

The integration of LLMs in peer review has evolved into two distinct approaches, each addressing specific needs in the review process. The first approach, \textbf{automated review generation}, emerged from the need to handle increasing submission volumes and reduce reviewer workload by using LLMs to analyze research papers independently \cite{Kousha2023ArtificialIT, Yu2024IsYP}. These systems are designed to evaluate multiple aspects of submissions, including methodology validation, results verification, and contribution assessment, thereby providing comprehensive review reports without direct human intervention. The second approach, \textbf{LLM-assisted review workflows}, developed in response to the recognition that human expertise remains crucial in academic evaluation while acknowledging that certain review tasks can benefit from automation\citep{Kuznetsov_2024_NLPForPeerReview}. These workflows incorporate LLMs as supplementary tools, where they assist human reviewers in time-consuming but well-defined tasks, such as paper summarization, reference verification, and internal consistency checks, while leaving critical evaluation and judgment to human experts.

These approaches employ diverse methodologies to enhance review efficiency, consistency, and quality. To systematically evaluate and improve these systems, the research community has developed specialized peer review benchmarks that serve dual purposes: providing standardized training datasets and establishing performance assessment metrics. This chapter examines these methodologies, their evaluation frameworks, and concludes with implementation challenges and future research directions.

\subsection{Automated Peer Review Generation}
\label{section6.2.1}

Automated peer review generation aims to streamline scientific assessment by exploring how LLMs can produce comprehensive reviews with minimal human intervention. By inputting a scientific article, these systems focus on generating a complete peer review or meta-review, employing various techniques to enhance feedback's depth, accuracy, and relevance.

Current approaches to automated peer review generation can be categorized into two main strategies: \textbf{single-model} and \textbf{multi-model} architectures. Single-model approaches focus on optimizing the review generation process through sophisticated prompting techniques and modular design. These systems typically employ carefully crafted prompts to guide the model's attention to specific aspects of the paper, such as methodology, results, and contributions \cite{Karmaker_2024_PromptingLLMsMetaReview}. 

Within the single-model paradigm, several distinct architectural approaches have been proposed.  {CGI2} \cite{zeng_2024_scientificopinionsummarizationpaper} advances beyond previous approaches: MetaGen \cite{Bhatia2020MetaGenAA}, which used a two-stage pipeline of extractive summarization with decision-aware refinement; Kumar et al. \cite{Kumar2021ADN}, which developed a neural architecture for joint decision prediction and review generation; and MReD \cite{Chenhui_2022_MReDDataset}, which introduced structure-controlled generation using sentence-level functional labels. Building on these foundations, CGI2 implements a staged review process through modular design, first extracting key opinions from the paper, then summarizing strengths and weaknesses, and finally refining these outputs through iterative feedback under a checklist-guided framework. This iterative process enhances the depth and relevance of reviews but may struggle with papers that involve highly complex methodologies or lengthy content exceeding the context window. Taking a different approach, {CycleReviewer} \cite{weng2024cycleresearcher} implements an end-to-end review generation method using reinforcement learning to refine review quality through feedback loops continuously. While CycleReviewer excels in enhancing review precision and clarity, its reliance on significant computational resources could limit its scalability. Meanwhile, {ReviewRobot} \cite{Wang_2020_ReviewRobot} utilizes knowledge graphs to systematically identify and structure knowledge elements, transforming them into detailed review comments through a structured generation process. ReviewRobot demonstrates remarkable explainability and evidence-based reasoning but is constrained by the inflexibility of its pre-defined templates.

The alternative strategy employs \textbf{multi-model} architectures, representing a more sophisticated approach by leveraging multiple specialized models to handle different aspects of the review process. This approach offers several advantages, including improved handling of complex papers and enhanced review quality through specialized expertise. {Reviewer2} \cite{gao2024reviewer2optimizingreviewgeneration} implements a two-stage process: one model generates specific aspect prompts, while another utilizes these prompts to create detailed, focused feedback. This separation of prompt generation and review creation allows for more nuanced and targeted feedback but often results in partial or biased reviews due to the lack of an integrated framework. To address these limitations, {SEA} \cite{Yu_2024_AutomatedPeerReviewing} employs separate models for standardization, evaluation, and analysis, providing a more comprehensive and balanced approach. The system unifies multiple reviews into a single format, significantly reducing redundancy and inconsistencies across feedback. Furthermore, SEA introduces a mismatch score to measure the alignment between papers and generated reviews, coupled with a self-correction strategy to enhance review quality iteratively. While these features enable SEA to surpass Reviewer2 in consistency and comprehensiveness, the need for coordinating outputs across multiple models introduces added complexity. Building on specialization but addressing a different challenge, {MARG} \cite{DArcy_2024_MARG} tackles the problem of processing papers that exceed typical LLM context limits. By introducing a multi-agent framework, MARG distributes review tasks across multiple specialized models, allowing for a comprehensive review of longer papers while maintaining attention to detail throughout the document. This innovative approach ensures detailed, aspect-specific feedback. Still, it brings new challenges, such as coordinating the communication and outputs of various agents, which increases the difficulty of ensuring consistency and alignment.

Each architectural approach offers distinct advantages and faces unique challenges. Single-model approaches benefit from simpler implementation and more straightforward control over the review process, but they may struggle with longer or more complex papers. Multi-model architectures provide greater scalability and better handling of sophisticated review tasks, yet they demand careful coordination and face potential consistency challenges across their components. For instance, ReviewRobot’s structured approach offers explainability and actionable insights. Still, it is less adaptable to evolving research domains, while CycleReviewer’s iterative refinement improves dynamic adaptability without requiring substantial training resources. As research in this area progresses, combining the strengths of single-model simplicity with the adaptability of multi-model designs presents a promising avenue for enhancing review quality, consistency, and comprehensiveness.

\subsection{LLM-assisted Peer Review Workflows}
\label{section6.2.2}
Unlike fully automated review generation, LLM-assisted peer review workflows focus on enhancing human reviewers' capabilities rather than replacing them. Recent research highlights the critical importance of this human-AI collaborative approach in academic peer review. Studies by \citep{Schintler_2023_EthicsAIPeerReview, Drori_2024_HumanInTheLoopAIReviewing, Biswas_2023_ChatGPTJournalReviews} emphasize that while LLM can enhance efficiency, human oversight remains essential for maintaining ethical standards and review integrity. Systems like AgentReview \citep{Jin_2024_AgentReview} demonstrate this synergy in practice, where LLM generates initial insights that human reviewers then refine and validate.

The LLM-assisted peer review workflows enhance three primary functions in the scientific review process: (1) information extraction and summarization, which helps reviewers quickly grasp paper content; (2) manuscript validation and quality assurance, which supports systematic verification of paper claims; and (3) review writing support, which assists in generating well-structured feedback.

In the information extraction and summarization function, systems automate document understanding and synthesis to support reviewer comprehension. PaperMage \citep{Lo_2023_PaperMageToolkit} is a foundational toolkit that integrates natural language processing and computer vision models to process visually rich scientific documents, enabling sophisticated extraction of logical structure, figures, and textual content across multiple modalities. Complementing this structural analysis, CocoSciSum \citep{Ding_2023_CocoSciSum} focuses on content summarization, offering customizable paper summaries with precise control over length and keyword inclusion while maintaining high factual accuracy through its compositional control architecture.

For the manuscript validation and quality assurance function, systems operate at different levels of analysis to ensure scientific rigor. At the local level, ReviewerGPT \citep{Liu_2023_ReviewerGPT} specializes in systematic error detection and guideline compliance, achieving high accuracy in verifying submission requirements while effectively identifying mathematical errors and conceptual inconsistencies within individual manuscripts. While ReviewerGPT focuses on internal manuscript validation, PaperQA2 \citep{Skarlinski_2024_LanguageAgentsSynthesis} performs global validation by examining claims against the broader scientific literature, employing sophisticated language agents to detect contradictions and verify assertions. The system demonstrates robust performance by identifying an average of 2.34 validated contradictions per paper while maintaining high factual accuracy in its cross-literature analyses. Additionally, Scideator \citep{Radensky2024ScideatorHS}, designed to facilitate idea validation, operates through facet recombination to identify novel and scientifically grounded analogies between papers. Scideator also includes a novelty checker, which evaluates claims for uniqueness and adherence to established research paradigms, offering reviewers enhanced capabilities to scrutinize manuscripts rigorously.

In the review writing support function, systems take different yet complementary approaches to assist reviewers at various expertise levels. ReviewFlow \citep{Sun_2024_ReviewFlowScaffolding} provides intelligent scaffolding through contextual reflection cues and note synthesis guidance, modeling expert practices to help novice reviewers produce well-structured reviews. The system's step-by-step approach benefits those new to peer review by decomposing the complex task into manageable components. 
While ReviewFlow focuses on individual reviewer guidance, CARE \citep{Zyska_2023_CAREEnvironment} emphasizes collaborative aspects of review writing through an integrated platform featuring NLP-enhanced inline annotations and real-time collaboration features, enabling reviewers to work together more effectively while providing detailed and constructive feedback \cite{DBLP:journals/tmlr/LiPD24, DBLP:conf/iclr/ChanCSYXZF024}. 
Further complementing these functionalities, DocPilot \citep{Mathur2024DocPilotCF} leverages modular task planning and code generation capabilities to automate repetitive and complex tasks in document workflows. Its structured approach to managing and annotating scientific PDFs ensures that reviewers can focus on substantive feedback rather than procedural hurdles, significantly improving their efficiency.

\subsection{Benchmarks}
\label{section6.3}
\begin{table*}[h!]
\centering
\caption{Peer Review Datasets and Evaluation Metrics. The Evaluation Metrics columns use the following abbreviations: PR (Peer Review), MR (Meta-review), S (Semantic Similarity), C (Coherence \& Relevance), D (Diversity \& Specificity), and H (Human Evaluation). Columns S, C, D, and H represent the evaluation metrics used in the study.}
\label{tab:peer_review_datasets}
\resizebox{\textwidth}{!}{
\begin{tabular}{>{\centering\arraybackslash}m{0.25\textwidth}|>{\centering\arraybackslash}m{0.05\textwidth}>{\centering\arraybackslash}m{0.05\textwidth}>{\centering\arraybackslash}m{0.6\textwidth}>{\centering\arraybackslash}m{0.05\textwidth}>{\centering\arraybackslash}m{0.05\textwidth}>{\centering\arraybackslash}m{0.05\textwidth}>{\centering\arraybackslash}m{0.05\textwidth}}
\toprule
\rowcolor{white}
\multirow{2}{*}{\textbf{Dataset Name}} & \multirow{2}{*}{\textbf{PR}} & \multirow{2}{*}{\textbf{MR}} & \multirow{2}{*}{\textbf{Additional Task}} & \multicolumn{4}{c}{\textbf{Evaluation Metrics}} \\
\cmidrule{5-8}
& & & & \textbf{S} & \textbf{C} & \textbf{D} & \textbf{H} \\
\midrule
MOPRD \citep{Lin_2023_MOPRD} & $\checkmark$ & $\checkmark$ & Editorial decision prediction, Scientometric analysis & $\checkmark$ & $\checkmark$ & $\checkmark$ & - \\
NLPEER \citep{Nils_2022_NLPeerResource} & $\checkmark$ & $\checkmark$ & Score prediction, Guided skimming, Pragmatic labeling & $\checkmark$ & $\checkmark$ & - & - \\
MReD \citep{Chenhui_2022_MReDDataset} & - & $\checkmark$ & Structured text summarization & $\checkmark$ & - & - & $\checkmark$ \\
PEERSUM \citep{Li_2023_SummarizingMultipleDocuments} & - & $\checkmark$ & Opinion synthesis & $\checkmark$ & $\checkmark$ & - & - \\
ORSUM \citep{zeng_2024_scientificopinionsummarizationpaper} & - & $\checkmark$ & Opinion summarization, Factual consistency analysis & $\checkmark$ & $\checkmark$ & - & $\checkmark$ \\
ASAP-Review \citep{yuan2021automatescientificreviewing} & $\checkmark$ & - & Aspect-level analysis, Acceptance prediction & $\checkmark$ & - & - & - \\
REVIEWER2 \citep{gao2024reviewer2optimizingreviewgeneration} & $\checkmark$ & - & Coverage \& specificity enhancement & $\checkmark$ & - & $\checkmark$ & - \\
PeerRead \citep{Kang_2018_PeerReadDataset} & $\checkmark$ & - & Acceptance prediction, Score prediction & $\checkmark$ & - & - & - \\
ReviewCritique \citep{Du_2024_LLMsAssistNLPResearchers} & $\checkmark$ & - & Deficiency identification & $\checkmark$ & - & $\checkmark$ & $\checkmark$ \\
\bottomrule
\end{tabular}}

\end{table*}

As automated review generation and LLM-assisted workflows continue to evolve, the research community faces a critical challenge: systematically evaluating and comparing these approaches. This need has led to the development of specialized benchmarks that assess various aspects of LLM-based peer review systems, from their ability to generate high-quality reviews to their effectiveness in supporting human reviewers.

The development and evaluation of LLM-based peer review systems rely on standardized benchmarks that assess different aspects of the review process. These benchmarks can be broadly categorized into three main types: 
(1) comprehensive review datasets that support holistic evaluation, including editorial decisions, scoring, and pragmatic analysis; 
(2) specialized assessment datasets that focus on specific aspects like opinion synthesis and consistency analysis;
and (3) quality evaluation datasets that measure review effectiveness through deficiency identification and acceptance prediction. Table~\ref{tab:peer_review_datasets} presents an overview of these key benchmarks and their associated evaluation frameworks.

The datasets, primarily sourced from publicly accessible academic conferences, serve diverse purposes in peer review tasks. Comprehensive datasets like MOPRD \citep{Lin_2023_MOPRD} and NLPeer \citep{Nils_2022_NLPeerResource} provide broad coverage, supporting tasks ranging from editorial decision prediction to pragmatic labeling. More specialized datasets focus on specific aspects of the review process: ASAP-Review \citep{yuan2021automatescientificreviewing} and Reviewer2 \citep{gao2024reviewer2optimizingreviewgeneration} emphasize acceptance prediction and coverage assessment. Recent additions such as ReviewCritique \citep{Du_2024_LLMsAssistNLPResearchers} introduce novel mechanisms for comparative analysis between human and LLM-generated reviews.

The evaluation framework for these benchmarks encompasses multiple dimensions, as detailed in Table~\ref{tab:peer_review_datasets}. Semantic Similarity measures how closely generated reviews align with reference texts, typically using metrics like ROUGE and BertScore. Coherence and Relevance evaluate the logical flow and topical appropriateness of reviews, while Diversity and Specificity assess the range and depth of feedback provided. Human Evaluation, incorporating expert assessment of review quality, offers crucial validation of automated metrics. Together, these four evaluation components - Semantic Similarity, Coherence and Relevance, Diversity and Specificity, and Human Evaluation - form a multi-faceted approach that ensures comprehensive assessment of LLM-generated reviews across various quality dimensions.

\subsection{Challenges and Future Work}

The integration of LLMs into academic peer review represents a significant shift in scholarly evaluation \cite{Liang_2024_MonitoringAIModifiedContent, liang2024mapping}. As academic institutions and publishers explore this technology, understanding its limitations and potential becomes crucial for the scholarly community.

\paragraph{Challenges} 
At the heart of peer review lies the need for deep expertise, nuanced understanding, and careful judgment. While LLMs show promise in supporting this process, their limitations reveal the complexity of automating scholarly assessment. A fundamental challenge is that LLMs often struggle to fully grasp the specialized terminology and complex concepts within academic fields. For example, in biochemistry, an LLM might misinterpret the significance of specific protein interactions, while in theoretical physics, it could fail to recognize subtle but critical assumptions in mathematical models \cite{Zhou_2024_LLMReliableReviewer}.

This limited technical comprehension directly impacts the LLM's ability to evaluate research methodology. When an LLM cannot fully understand field-specific concepts, it cannot reliably assess whether the research methods are appropriate or if the evidence justifies the conclusions. For instance, where methodological standards vary across fields in interdisciplinary research, LLMs often fail to identify critical issues such as inadequate sample sizes, inappropriate statistical tests, or missing experimental controls \cite{Robertson_2023_GPT4PeerReviewAssistance}. This limitation becomes particularly concerning given the high stakes of peer review in ensuring research quality and scientific integrity.

The complexity of academic writing introduces additional challenges, mainly when dealing with longer manuscripts. Even as context windows expand, LLMs struggle to maintain coherent analysis across extensive texts, often losing track of complex arguments spanning multiple sections. This limitation frequently results in inconsistent or contradictory evaluations \cite{Chamoun_2024_AutomatedFocusedFeedback}. More concerning is the persistent issue of hallucination—models sometimes generate convincing but incorrect assessments, particularly when reviewing novel research approaches \cite{DArcy_2024_MARG}.

Furthermore, implementing LLMs in peer review faces additional challenges beyond technical performance limitations. A fundamental infrastructure issue is the shortage of specialized training data \cite{zeng_2024_scientificopinionsummarizationpaper, Kang_2018_PeerReadDataset}, which creates an uneven landscape across academic disciplines. This data scarcity particularly affects fields with smaller research communities or specialized vocabularies. Equally concerning are the ethical implications of LLM-assisted peer review. Issues of algorithmic bias and transparency \cite{Schintler_2023_EthicsAIPeerReview} have emerged alongside new forms of academic misconduct, such as ``plagiarism laundering'' \cite{Piniewski_2024_EmergingPlagiarismPeerReview}. Additionally, a critical concern is the potential homogenization of academic feedback if many researchers rely on the same LLM systems for peer review~\cite{Liang_2024_MonitoringAIModifiedContent}. The widespread use of similar AI tools may reduce the diversity of perspectives and diminish the creative insights that stem from the distinct thought processes of individual human reviewers.

\paragraph{Future Work}
To advance LLMs' capabilities in academic paper reviewing, several fundamental technical challenges must be prioritized. First, current LLMs struggle with specialized technical concepts across different academic fields, necessitating improved approaches for processing and understanding domain-specific terminology. Second, we need enhanced citation analysis capabilities to verify reference relevance and assess how effectively citations support a paper's arguments. Third, analyzing long academic documents requires new methods for maintaining coherence - from cross-referencing between sections to verifying consistency across methods, results, and conclusions.

Beyond technical improvements, developing effective human-AI collaboration frameworks is crucial. The next generation of review systems must create intuitive interfaces that highlight potential issues and integrate seamlessly with human workflow \cite{Drori_2024_HumanInTheLoopAIReviewing}. These collaborative systems must be adaptable across different academic fields, with special consideration for disciplines with limited computational resources \cite{Karmaker_2024_PromptingLLMsMetaReview}. Rigorous evaluation frameworks for these human-AI systems must ensure they genuinely enhance reviewer efficiency and effectiveness \cite{liiqa, DBLP:conf/chi/XiaoDLEK0L24}.

As LLM becomes more prevalent in peer review, robust governance mechanisms become critical. This includes developing reliable methods for detecting LLM-generated content, ensuring transparent tracking of LLM contributions, and maintaining reviewer authenticity \cite{Liang_2024_MonitoringAIModifiedContent}. Additionally, we need standardized protocols for securely integrating LLM review tools with existing journal platforms \cite{Altmäe_2023_AIScientificWriting}.

Lastly, progress in these areas must be measured through comprehensive evaluation frameworks. For technical capabilities, we need systematic assessments of improvements in language understanding, citation analysis, and document coherence. Human-AI collaboration metrics should evaluate the quality of LLM suggestions and their impact on reviewer efficiency. Governance evaluations must assess the reliability of LLM detection systems and the security of platform integrations. Crucially, these frameworks should examine potential biases across different academic disciplines, publication formats, and linguistic backgrounds to ensure equitable support for all scholarly communities. Through these targeted assessments, we can guide the development of LLM systems that meaningfully enhance the peer review process while maintaining its integrity.

\section{Conclusion}

This survey comprehensively explores the transformative role of LLMs throughout the scientific research lifecycle, from hypothesis generation and experiment to writing and peer reviewing. By identifying both the opportunities and challenges in applying LLMs to these tasks, we highlight their current capabilities, limitations, and potential to enhance scientific productivity. In conclusion, LLMs represent advanced productivity tools, offering new methods across all stages of modern scientific research. Despite being constrained by inherent limitations, technical barriers, and ethical considerations in domain-specific tasks, the continued advancement of LLM capabilities promises to revolutionize research practices. As these systems evolve, their integration into scientific workflows will not only accelerate discoveries but also foster unprecedented innovation and collaboration in the scientific community.



\section*{Limitations}

The general concept ``AI for Science'' is a huge topic, and this survey only focuses on the LLMs for scientific research aspect.
In addition, many researchers from ``science'' background but not computer science background might also have conducted works in this domain, but might not published in a computer science venues.
We might have missed some of these works in this survey.




\bibliography{custom, 5_reviewing, implementation, discovery, paper_writing}
\bibliographystyle{ACM-Reference-Format}


\end{document}